\documentclass[review]{elsarticle}

\usepackage{lineno,hyperref}
\modulolinenumbers[5]

\journal{Robotics and Autonomous Systems}

\usepackage{amsmath}
\usepackage{amssymb}
\usepackage{booktabs}
\usepackage{graphicx}
\usepackage{subfigure}
\usepackage[font={footnotesize},labelfont=bf]{caption}
\usepackage{microtype}
\usepackage{textcomp}
\usepackage[utf8]{inputenc}
\usepackage{hyperref}
\usepackage{xcolor}

\usepackage[small,compact]{titlesec}
\usepackage{lscape}
\usepackage[margin=3.3cm]{geometry}%

\usepackage[font=small,labelfont=bf]{caption}
\usepackage{booktabs}

\begin{document}

\begin{frontmatter}

\title{Learning Whole-Image Descriptors for Real-time Loop Detection and Kidnap Recovery under Large Viewpoint
Difference}
\tnotetext[mytitlenote]{The source code is available at \url{https://github.com/HKUST-Aerial-Robotics/VINS-kidnap} }

\author{Manohar Kuse\fnref{myfootnote}}
\author{Shaojie Shen}
\address{Robotics Institute, Hong Kong University of Science and Technology, Clear Water Bay Road, Kowloon, Hong Kong}
\fntext[myfootnote]{Corresponding author: mpkuse@connect.ust.hk}

\begin{abstract}
    We present a real-time stereo visual-inertial-SLAM system
    which is able to recover from complicated kidnap scenarios and failures online in realtime.
    We propose to learn the whole-image-descriptor in a weakly supervised manner based on
    NetVLAD and decoupled convolutions.
    We analyse the training difficulties in using standard loss formulations and propose an allpairloss
    and show its effect through extensive experiments.
    Compared to standard NetVLAD, our network takes an order
    of magnitude fewer computations and model parameters, as a result runs about three times faster.
    We evaluate the representation power
    of our descriptor on standard datasets with precision-recall.
    Unlike previous loop detection methods which have been
    evaluated only on fronto-parallel revisits,
    we evaluate the performace of our method with competing methods
    on scenarios involving large viewpoint difference. Finally, we present the fully functional
    system with relative computation and handling of multiple world co-ordinate system
    which is able to reduce odometry drift, recover from complicated kidnap scenarios and
    random odometry failures.
    We open source our fully functional system
    as an add-on for the popular VINS-Fusion.
\end{abstract}


\end{frontmatter}

\section{Introduction}

\begin{figure}[ht]
    \centering
        \includegraphics[width=0.95\columnwidth]{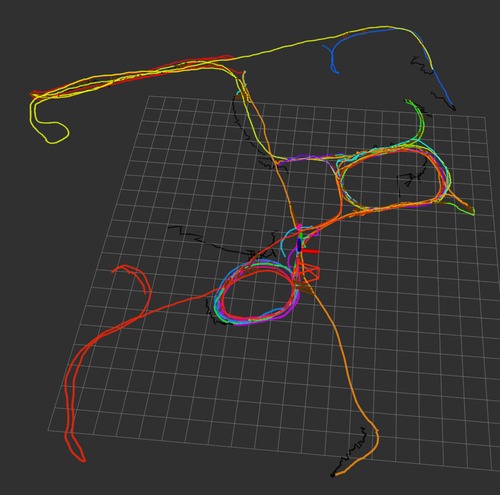}
    \caption{Shows the corrected trajectories (different colors for different worlds) merged
    according to the inter-world loop candidates. Note that the merging
    occurs live (not offline) in real-time as the loop candidates are found. We also note
    that such cases cannot be handled by Qin \textit{et al.}\cite{qin2018relocalization} which
    just merges with the world-0 (first world) and
    ignore any inter-world loop candidates not involving world-0.
    The \textit{maplab} system \cite{schneider2018maplab} provides an online tool,
    \textit{ROVIOLI} which is essentially
    a visual-inertial odometry and localization front-end. Although it provides
    for a console based interface offline for multi-session map merging, it cannot identify kidnaps
    and recover from them online.
    This sequence
    involves multiple kidnaps lasting from 10s to 30s.
    The video for the live run is available at the link: \url{https://youtu.be/3YQF4_v7AEg}.
    Live runs videos are available for more
    sequences through this link: \url{https://bit.ly/2IkEh3F}. }
    \label{fig:kidnap-screenshot}
    \vspace{-.25in}
\end{figure}

Over the past decade, the
SLAM (Simultaneous Localization and Mapping ) community has made amazing progress towards increasing the specificity
of the odometer and building usable maps of the environment to assist robots in
various planning tasks. Systems using visual and inertial
information fusion have been a contemporary theme towards reducing drift to less than 0.5\% of the trajectory length
 \cite{cadena2016past}.
Identifying a revisit to a place presents an
opportunity to reduce the drift
further and also to recover from kidnap scenarios.
General place recognition, however, remains an extremely challenging problem \cite{lowry2016visual} due to
myriad ways in which visual appearance of a place varies.
In our own daily experience, humans describe places to
fellow humans as a collection of objects,
their color cues, their spatial locations and so on, thereby
allowing to disambiguate places even
if they approach the place from a very different viewpoint.
Ideally, the loop detection module should describe a scene in this context.
Humans probably do not rely on corner features (a
common technique in use for loop detection in existing SLAM systems)
to identify a place,
instead we humans, most likely represent the scene as a whole in a semantic sense.
The proposed system builds on this motivation.

In this work, we propose the use of a framework which learns whole-image
descriptors without explicit human labeling
to represent a scene in a high dimensional subspace for detecting
place revisits. We lay special emphasis on the
real-time performance and evaluation of the system
in context of visual-SLAM.

Popular past works
 have considered loopclosure under fronto-parallel
scenarios. However,
place revisits can happen at substantial viewpoint difference.
The underlying place recognition module in
 SLAM systems to identify place revisits
occurring at widely different viewpoints.
Past systems based on bag-of-visual-words (BOVW) are limited by the underlying low-level feature
descriptors. The learned vocabulary (for BOVW) also have difficulty
generalizing under adversaries like
large viewpoint difference, noise, low light, changing exposure, less texture
\cite{lowry2016visual}.
The proposed method can learn a representation that generalizes well
and can identify place revisits under non fronto-parallel viewpoints.

We compare our method's run-time performance with a popular
bag-of-words approach, DBOW \cite{galvez2012dbow} and
ibow-lcd by \cite{garcia2018ibow}  along with recently proposed CNN based approaches for
place recognition \cite{sunderhauf2015performance}, \cite{merrill2018lightweight}, \cite{antequera2017}.
On real sequences, our method
delivers a similar recognition performance to NetVLAD but at a 3X lower
computational time and an order of magnitude fewer
training variables.
The major advantage of our system is that it has a high place recall rate
thus is able to recover live in real-time
from long and chained kidnaps by maintaining multiple co-ordinate systems and their relative
poses.

Our paper is organized as follows. In Section \ref{sec:lit-review}, we start by reviewing
approaches in the
Visual Place Recognition (VPR) community and some recent loopclosure methods used in
Visual-SLAM community. Next, in Section \ref{sec:learning_core}, we identify the
issue of unstable learning
in the original NetVLAD implementation which uses the tripletloss and we
propose an allpairloss function to alleviate this issue.
In Section \ref{sec:desc_extraction_comparison} we present our implementation details for
deployment as a visual-SLAM subsystem which includes the
place recognition module, the datastructure for handling multiple co-ordinate systems
and recovery from kidnap.
In Section \ref{sec:all_experiments},
we presents comparative experiments to this effect. Finally, we present our entire
system which is available as a pluggable module to the popular VINS-Fusion  \cite{vins-mono}.

\section{Literature Review}
\label{sec:lit-review}
We recognize that visual place recognition (VPR) and loopclosure
detection in SLAM are related problems. Here we first review
recent advances from VPR community and then review state-of-the-art
loop-closure methods.

In the context of VPR, Sunderhauf \textit{et al.} \cite{sunderhauf2015performance} pioneered
the use of ConvNet features. Compared to SeqSLAM \cite{milford2012seqslam}
and FAB-MAP \cite{cummins2008fab,cummins2011appearance} use of features from pretained network results
 in better precision-recall performances on standard VPR datasets(Norland, Gardens Point, St. Lucia and Campus).
In their subsequent work, Sunderhauf \textit{et al.} \cite{sunderhauf2015place} proposed to use region proposals
and extract ConvNet features on each of the regions.
Arandjelovic \textit{et al.} \cite{arandjelovic2016netvlad}
proposed a trainable feature aggregation layer which mimics the popular VLAD (Vector of Locally Aggregated
Descriptor). While impressive performance was obtained, these methods rely on
nearest neighbour search for retrival. The image descriptor being very high dimensional (eg. 32K dimensional for
\cite{arandjelovic2016netvlad}, 64K for \cite{sunderhauf2015performance}), these methods perform various dimensionality reduction techniques to make
nearest neighbour search feasible in reasonable time with some hit to the retrieval performance. WPCA was used by \cite{arandjelovic2016netvlad} which involve storage of a 32Kx4K
matrix costing about 400 MB.

More recently Khaliq \textit{et al.} \cite{khaliq2018holistic} proposed an
approach which make use of region-based features from a light-weight CNN architecture and combines them with VLAD aggregation.
The approach from Chen \textit{et al.} \cite{chen2018learning,chen2017only}
identifies key landmark regions directly from responses of VGG16 network
which was pretrained on image classification task.
For regional features encoding,
bag-of-words was employed on a separare training dataset to learn the codebook.
The approach by Hou \textit{et al.} \cite{hou2018bocnf} is very similar to \cite{chen2018learning}.
The Disadvantage of using pretrained models learned on ImageNet
 object classification, for
 example, puts more emphasis on objects rather than the nature
 of the scene itself.
Other works in this context include \cite{bai2018sequence,gao2017unsupervised,7298790,7989366,7989359,babenko2015aggregating,arandjelovic2014dislocation,sattler2016large}.
For a more detailed summary of the works in place recognition
we direct the readers to survey on place recognition / instance
retrieval \cite{zheng2018sift,lowry2016visual}. We summarize
the literature in Table \ref{tab:place-recog-survey}.

Although CNN based techniques are considered as state-of-the-art
in retrieval and place recognition tasks, they are still disconnected from overall SLAM
and loop-closure detection problems. Commonly employed loop detection methods in
state-of-the art SLAM systems rely on sparse point feature descriptors like SIFT, SURF, ORB, BRIEF etc.
for representation and an adaptation of BoVW for retrieval.
While BoVW provides for an scalable indexed retrieval,
the performance of the system is limited by the underlying
image representation. Such factors as the quantization in
clustering when building vocabulary, occlusions,
image noise, repeated structures also affect the retrieval performance.

FAB-MAP \cite{cummins2008fab,cummins2011appearance}, DBOW2 \cite{galvez2012dbow} and
others \cite{mur2014fast,bampis2017high}  rely on a visual vocabulary which is trained offline,
while recent methods like OVV \cite{nicosevici2012automatic} , IBuILD\cite{khan2015ibuild},
iBOW-LCD \cite{garcia2018ibow}, RTAB-MAP \cite{labbe2013appearance}
 and others \cite{angeli2008fast,garcia2014use,zhang2016learning,stumm2016_locationmodels} rely on online constructed visual vocabulary.
Authors have also made use of whole-image-descriptors in loopclosure context \cite{Zhang2016RobustMS,pepperell2014_allenv,milford2012seqslam}.
Works in the context of loopclosures in SLAM which
make use of learned feature descriptors are:
\cite{merrill2018lightweight, gao2017unsupervised}.
Merril and Huang \cite{merrill2018lightweight} learned
an auto-encoder from
the common HOG descriptors for the whole image.
Other miscellaneous work related to our localization system are
\cite{kenshimov2017deep,DBLP:journals/corr/FeiTS15,sizikova_eccvworkshop2016,cieslewski2017_distributed,cieslewski2017_wholeimagedesc}.

Some works have also built full SLAM system with multi-session map merging capability.
The \textit{maplab} system \cite{schneider2018maplab} provides an online tool,
\textit{ROVIOLI} which is essentially
a visual-inertial odometry and localization front-end. Although it provides
for a console based interface offline for multi-session map merging, it cannot identify kidnaps
and recover from them online. This is the major distinguishing point of our system.
Also the relocalization system by Tong \textit{et al.} \cite{qin2018relocalization}
can merge multiple sessions live it only merges with the first co-ordinate frame, any loop
connections between co-ordinate systems not involving the first co-ordinate systems
are ignored. Our system on the other hand is able to maintain multiple co-ordinate systems
and their relative poses, set associations and merge the trajactories online in real-time.

We summarize our contributions:
\begin{itemize}
\item A fully functional system, as an add-on for VINS-Fusion, which uses whole-image descriptor
for place representation and recovery from odometry drifts, kidnap and failures live
and in real-time.
Our learning code \footnote{\url{https://github.com/mpkuse/cartwheel_train}} and
VINS-Fusion addon ROS package \footnote{\url{https://github.com/HKUST-Aerial-Robotics/VINS-kidnap}} are open sourced.
\item A novel cost function which deals with the gradient
issue observed in standard NetVLAD training.
\item Decoupled convolutions instead of standard convolutions
result in similar performance on precision-recall basis
but at a 3X lower computation cost and about 5-7X fewer
learnable parameters, making it ideally suitated for real-time
loopclosure problems.
\item Squashing channels of CNN descriptors
instead of explicit dimensionality reduction of image
descriptor for scalabilty. Even a 512-D image descriptor
gives reasonable performance.
\end{itemize}

\begin{landscape}
\begin{table*}[]
\begin{tabular}{|p{4cm}|l|p{10.5cm}|}
\toprule
\textbf{Representation} & \textbf{Retrieval} & \textbf{Method Description} \\
\midrule
SPF-real                & BOW                & \cite{cummins2008fab, cummins2011appearance} soft-real-time (can run @5-10hz) \\
SPF-binary              & BOW                & \cite{galvez2012dbow, mur2014fast,bampis2017high} real-time  (10hz or more)\\
SPF-real                & Inc-BOW            & \cite{angeli2008fast,nicosevici2012automatic,labbe2013appearance}
		soft-real-time (1-5 Hz). \cite{tsintotas2018assigning, bampis2018fast}   make use of temporal information to form visual words scene representation.\\
SPF-binary              & Inc-BOW            & \cite{garcia2014use,zhang2016learning,khan2015ibuild,garcia2018ibow} soft-real-time to 1-5Hz 					processing  \\
SPF                     & graph              & \cite{stumm2016robust} \\
Pretrained-CNN          & NN                 & \cite{sunderhauf2015performance,arroyo2016fusion,bai2018sequence}
			provide for real-time descriptor computation (10-15hz). dimensionality reduction accomplished at 5hz, NN with 64K dim is really slow, NN after dimemsionality reduction (~4000d)  is about 			5-15 hz.  \\
Pretrained-CNN          & BOW                & \cite{hou2018bocnf,chen2018learning,chen2017only} \\
Custom-CNN       & NN                 & \cite{arandjelovic2016netvlad,7298790,7989366,7989359} provides for real-time descriptor computation. dim-reduction 					and NN search are bottle necks. \\
Custom-CNN with region-proposals & regionwise-NN & \cite{sunderhauf2015place} very slow representation vector computation.
  \cite{khaliq2018holistic} region descriptor encoding computation 2-3 Hz. Reported matching times is several seconds. \\
Unsupervised Learning   & NN                 & \cite{merrill2018lightweight,gao2017unsupervised,antequera2017} descriptors are not descriptive enough after 					dim-reduction. real-time desc computation.  \\
Intensity               & NN                 & \cite{milford2012seqslam,pepperell2014_allenv} \\
agnostic                & optimization       & \cite{Zhang2016RobustMS,latif2014online,han2018sequence} generally slow. \\
\bottomrule
\end{tabular}
\caption{A summary of visual place recognition literature. SPF=Sparse Point Features. BOW=Bag-of-words.}
\label{tab:place-recog-survey}
\end{table*}
\end{landscape}

\section{Learning Place Representation}
\label{sec:learning}
\label{sec:learning_core}
In this section, we describe the training procedure.
We start by reviewing VLAD and NetVLAD \cite{arandjelovic2016netvlad}
(Sec. \ref{sec:netvlad}). We see these methods as a way for pixelwise featuremap
aggregation. Next we describe the learning issues
associated with the triplet ranking loss function.
To mitigate this issue we propose to use a novel all-pair loss function (Sec. \ref{sec:cost_function}).
We provide an  intuitive explanation along with experimental
evidence on why the proposed all-pair loss function leads to faster
and stable training.

\subsection{Review of VLAD and NetVLAD}
\label{sec:netvlad}

\begin{figure}[]
\centering
\begin{subfigure}
\centering
\includegraphics[width=0.42\textwidth]{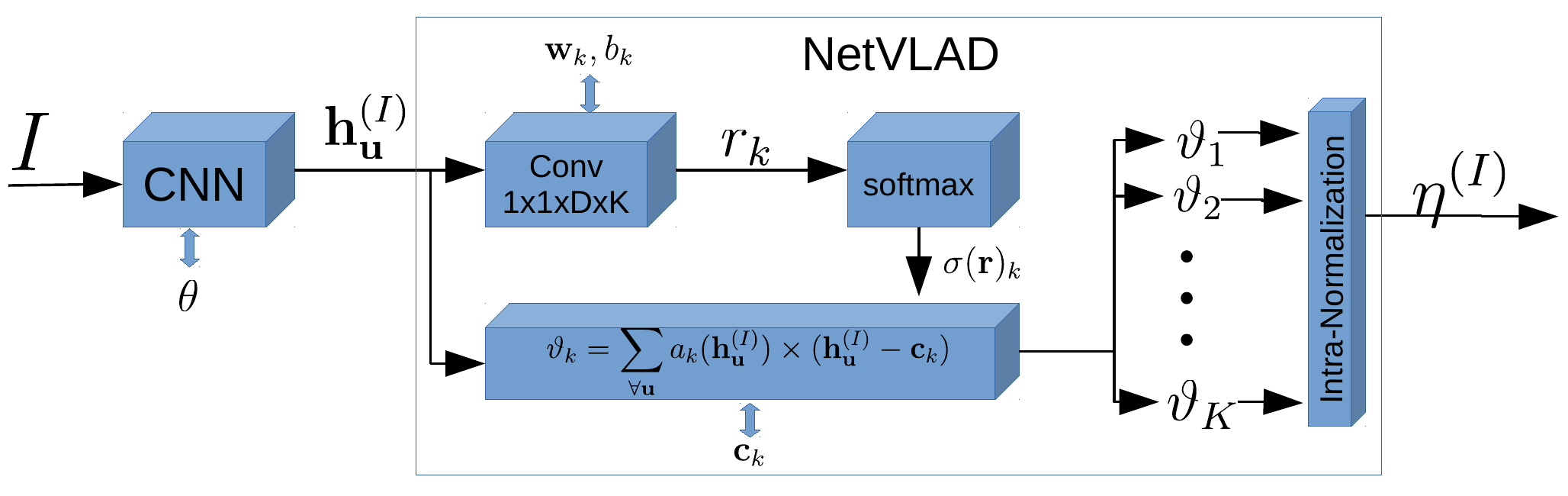}
\end{subfigure}
\caption{Notations and computations for the whole-image descriptor.
An image is fed into the CNN followed by the NetVLAD layer.
We experiment with VGG16 and propose to use
decoupled convolution for its speed. Additionally
for dimensionality reduction we propose channel-squashing.
Our fully convolutional network, with K=16 produces a
4096-dimensional image descriptor (without channel squashing)
and a 512-dimensional image descriptor (with channel squashing).
In terms of number of floating point operations (FLOPs) for a 640x480
input image our proposed network is about 25X faster, real
computational time is about 3X faster.
Details in Sec. \ref{sec:learning_core}.
}
\label{fig:cnn_plus_netvlad}
\end{figure}

Let ${\bf{h}}^{(I)}_{\bf{u}}(d)$ be the $d^{th}$ dimension ($d = 1 \cdots D$)
of the image feature descriptors for image $I$ (width $W$ and
height $H$) at pixel $\mathbf{u} := (i,j), i=1,\ldots W' ; j=1,\ldots H'$.
These per pixel CNN-feature descriptors  are assigned to one of the K clusters (K is a
fixed parameter, we used 16, 48, 64 in our experiments),
with ${{\bf{c}}_k} \in \Re^{D}, k=1 \ldots K$ as cluster centers.
The VLAD \cite{arandjelovic2016netvlad} representation is $D \times K$ matrix,
$\mathbf{V} = [ \mathbf{\vartheta}_1, \mathbf{\vartheta}_2, \ldots, \mathbf{\vartheta}_K ]$,
defined as the sum of difference between local descriptor and assigned cluster center,

\begin{equation}
 {\mathbf{\vartheta}}_k = \sum_{\forall \mathbf{u}} a_k( {\mathbf{h}}^{(I)}_{\mathbf{u}} ) \times ( {\mathbf{h}}^{(I)}_{\mathbf{u}} - {\mathbf{c}}_k )
\end{equation}

where $a_k(.)$ denotes a scalar membership indicator function of the descriptor
${\mathbf{h}}^{(I)}_{\mathbf{u}}$
in one of the K classes.
Arandjelovic \textit{et al.} \cite{arandjelovic2016netvlad} proposed to mimic VLAD in a CNN-based framework.
In order that the cluster assignment function, $a_k(.)$, be differentiable and hence learnable with
back propagation they defined an approximation of the assignment function $a_k(.)$
using the softmax function. For brevity,
we write, ${\mathbf{h}}^{(I)}_{\mathbf{u}}$ as $\mathbf{h}$:

\begin{equation}
\label{eq:cluster_association}
\begin{split}
 \hat{a}_k( \mathbf{h} ) &= \frac{e^{-\alpha \ || \mathbf{h} - \mathbf{c}_k || }}{ \sum_{k'=1}^{K} e^{-\alpha \ || \mathbf{h} - \mathbf{c}_{k'} ||} }\\
	  &= \sigma( \mathbf{r} )_k
\end{split}
\end{equation}

where $r_k=\mathbf{w}_k^{T} \mathbf{h} + b_k$,
$\mathbf{w}_k = 2\alpha \mathbf{c}_k$, $b_k = -\alpha ||\mathbf{c}_k||^2$.
$\sigma( \mathbf{r} )_k$ is the softmax function. $r_k$ can be computed with convolutions.
$\mathbf{w}_k$, $b_k$ and $\mathbf{c}_k$ are learnable parameters
in addition to the CNN parameters $\theta$.
Fig. \ref{fig:cnn_plus_netvlad} summarizes the computations and notations.
Each of the vectors corresponding to K clusters is individually
unit normalized and then the whole vector is unit normalized. This is
referred in the literature as Intra-normalization which reduce the effect of burstiness
of visual features\cite{jegou2009burstiness}.
Thus, scene descriptor $\mathbf{\eta}^{(I)}$, of size
$D*K$ is produced using a CNN and the NetVLAD layer,
\begin{equation}
 \mathbf{\eta}^{(I)} = \aleph( \{ {\mathbf{h}}^{(I)}_{\mathbf{u}} \} )
\end{equation}

In general any base CNN can be used and the NetVLAD
mechanism can be thought of  aggregating the
CNN pixel-wise descriptors. For experiments, we
use the VGG16 network \cite{simonyan2014very}.
Additionally, following \cite{howard2017mobilenets} we
propose to use the decoupled convolution, ie.
a convolution layer is split into two layers, the first of which
does only spatial convolution independently across all the
input channels.
Second of the two layers does 1x1 convolution
on the channels. This has been found to boost running time
at marginal loss of accuracy for object categorization tasks.
We also propose to reduce the dimensions by quashing
channels with learned 1x1 convolutions rather than
reduce the dimensions of the image descriptor as has been
done by the original NetVLAD paper. This eliminates the need to store
the whittening matrix (as done by \cite{arandjelovic2016netvlad}). At the
run time it eliminates the need for a large matrix-vector multiplication for dimensionality
reduction.

\subsection{Proposed All-Pair Loss Function}
\label{sec:cost_function}
To learn the parameters of the CNN ($\theta$) and of the NetVLAD layer
($\mathbf{w}_k$, $b_k$ and $\mathbf{c}_k$),
the cost function needs to be designed such that, in the dot product space,
$\eta$ corresponding to projections of the same scene (under different viewpoints) appear as nearby points (higher dot product value, nearer to 1.0).
Let $\eta^{(I_q)}$ be the descriptor of the query image $I_q$. Similarly,
let $\eta^{(P_i)}$ and $\eta^{(N_j)}$ be the descriptors of $i^{th}$ positive and $j^{th}$ negative sample respectively. By positive sample, we refer to a scene which is same as query
image scene but imaged from a different perspective. By negative sample, we
refer to a scene which is not the same place as the query image.
Let the notation, $\langle \eta^{(a)}, \eta^{(b)}\rangle$, denote
the dot product of two vectors.

Following \cite{arandjelovic2016netvlad} we use multiple
positive and negative samples  $\{I_q, \{P_i\}_{i=1,\ldots,m}, \{N_j\}_{j=1,\ldots,n}\}$ per training sample, however with a novel
all-pair loss function.
We provide an  intuitive  explanation  for the superiority of the proposed
loss function over the standard triplet loss  used by  \cite{arandjelovic2016netvlad} for training.
We also provide corroborative experimental evidence towards our claims.
The commonly used triplet loss function
can be rewritten in our notations as, $L_{triplet-loss}$:
\begin{equation}
\label{eq:netvlad_original}
 \sum_j \mbox{max}\big( 0, \langle \eta^{(I_q)}, \eta^{(N_j)}\rangle - \mbox{min}_i ( \langle \eta^{(I_q)}, \eta^{(P_i)} \rangle ) + \epsilon \big)
\end{equation}

where $\epsilon$ is a constant margin. Note that \cite{arandjelovic2016netvlad}
prefered to define the loss function in Euclidean space rather than
the dot product space. Using any of the spaces is equivalent
since for unit vectors,
$\bf{a}$ and $\bf{b}$, the dot product, $\langle \bf{a}, \bf{b} \rangle$,
and the squared Euclidean distance $d(\bf{a},\bf{b})$, are related as,
$d(a,b) = 2 (1 - \langle a, b \rangle)$ with a negative correlation.
This has been taken care of by flipped sign in our optimization
problem compared to the one used by Arandjelovic \textit{et al.} \cite{arandjelovic2016netvlad}.
This loss function is
the difference between the worst
positive sample, ie. $\mbox{min}_i \langle \eta^{(I_q)}, \eta^{(P_i)} \rangle$
and the query with every negative sample.

In an independent study by Bengio \textit{et al.} \cite{bengio2009curriculum}, it was observed that
for faster convergence,
it is crucial to select triplets from the training dataset,
that violate the triplet constraint, ie. result in
as few zero-loss as possible. They demonstrated that these zero-loss
scenarios lead to zero gradients which in turn slows the training.
They suggested to provide easier samples in early iterations and
harder samples in later iteration to speed up the learning process.
To this effect,
Schroff \textit{et al.} \cite{schroff2015facenet} proposed
a strategy to select triplets using recent network checkpoints, every
$n$ (say 1000) training iterations.
Instead of using a complicated strategy as done by \cite{schroff2015facenet},
we rely on a well-designed loss function which gives this effect.
Thus, we define a novel loss function based on all-pair comparisons of
positive and negative samples with the query image.
The proposed loss function is
relatively harder to satisfy (resulting in fewer zero loss samples), hence its higher discriminatory power compared to the
triplet loss (see Fig. \ref{fig:batch-zero-loss-compare}).

\begin{figure}
\centering
 \includegraphics[scale=.4,angle=90]{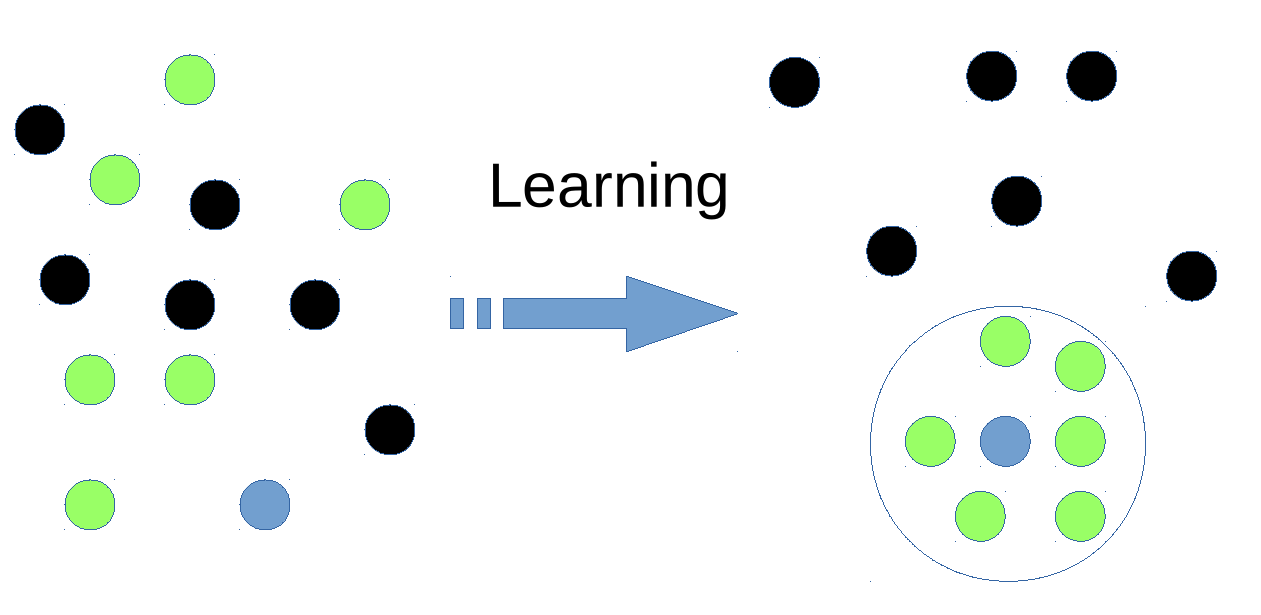}
 \caption{Illustration of the effect of learning with proposed loss function.
 Descriptor of query image ($\eta^{(I_q)}$, in blue). Descriptors of
 positive set ($\eta^{(P_i)}$, in green) and negative set ($\eta^{(N_j)}$, in black). See also Eq. \ref{eq:cost_final}.}
 \label{fig:cost_function_schematic}
\end{figure}

In order to learn highly
discriminative descriptors,
we want the similarity of query sample ie. $\eta^{(I_q)}$ with positive samples be more than
the similarity between query sample and the negative samples. Let us consider two cases
a) $\langle \eta^{(I_q)}, \eta^{(N_j)} \rangle > \langle \eta^{(I_q)}, \eta^{(P_i)} \rangle$;
b) $\langle \eta^{(I_q)}, \eta^{(N_j)} \rangle < \langle \eta^{(I_q)}, \eta^{(P_i)} \rangle$.
Case-b is what we prefer so we do not want to have a penalty (want to have zero loss) for its occurence.
 Case-a is opposite of
what we prefer thus we add a penalty proportional to the magnitude of the dot product to discourage this event.
We propose to add a penalty term as above for all the pairs where the
conditions do not hold.
For effective learning we propose to compute the loss over every
pair of positive and negative sample. The final loss function for one
learning sample ( $\{I_q, \{P_i\}_{i=1,\ldots,m}, \{N_j\}_{j=1,\ldots,n}\}$  ) is given
as, $L_{proposed}$:

\begin{equation}
\label{eq:cost_final}
 L = \sum_{i=1}^{m} \sum_{j=1}^{n} \mbox{max}( 0, \langle \eta^{(I_q)}, \eta^{(N_j)} \rangle - \langle \eta^{(I_q)}, \eta^{(P_i)} \rangle + \epsilon )
\end{equation}
The motivation and the effect of the proposed loss function on
termination of learning is summarized
in Fig. \ref{fig:cost_function_schematic}.
We further note that the proposed loss function is harder to
satisfy (giving a positive penalty) compared to the
loss function used by Arandjelovic \textit{et al.}\cite{arandjelovic2016netvlad} (ie. Eq. \ref{eq:netvlad_original}). This has been experimentally observed (see Fig. \ref{fig:batch-zero-loss-compare}).
The major drawback of using an easy to satisfy penalty function
is the vanishing gradients problem \cite{bengio2009curriculum},
which slows the speed of learning. This is because a zero-loss sample
results in zero-gradient during back-propagation.

\subsubsection{Loss function in Matrix Notation}
\label{sec:matrix_notation}
For fast and efficient training we express the loss function (Eq. \ref{eq:cost_final}) in a matrix notation.
We firstly define $\varDelta^{q}_{\bf{P}}$ to represent dot product between the sample query $\eta_q$ and descriptors of
each of the positive samples

\begin{equation}
 \varDelta^{q}_{\bf{P}} =
      \begin{bmatrix} \langle \eta^{(I_q)}, \eta^{(P_1)} \rangle \\ \vdots  \\ \langle \eta^{(I_q)}, \eta^{(P_m)} \rangle  \end{bmatrix}
\end{equation}

Next, we define $\varDelta^{q}_{\bf{N}}$ to represent dot product between the sample query and descriptors of each of the
negative samples.

\begin{equation}
 \varDelta^{q}_{\bf{N}} =
      \begin{bmatrix} \langle \eta^{(I_q)}, \eta^{(N_1)} \rangle \\ \vdots  \\ \langle \eta^{(I_q)}, \eta^{(N_n)} \rangle  \end{bmatrix}
\end{equation}

Let $\mathbf{1}_n$ denote a column-vector of size $n$ and $\mathbf{1}_m$
denote a column-vector of size $m$ with all entries as 1s.
Also define $\mathbf{0}_{m \times n}$ as null-matrix of dimensions $m \times n$.
The max(.) operator is a point-wise operator.
Now
we note that Eq. \ref{eq:cost_final} can be expressed in matrix notation as:
\begin{equation}
 \mathbf{L} =  \mbox{ max }( \mathbf{0}_{m \times n},  \mathbf{1}_m (\varDelta^{q}_\mathbf{N})^T  -   \varDelta^{q}_{\mathbf{P}} \mathbf{1}_n^T +  \epsilon \mathbf{1}_m \mathbf{1}_n^T )
\end{equation}

\subsection{Training Data}

In order to train the scene descriptor, only requirement on the data is that
we be able to draw positive sample images (views of the same physical scenes)
and  negative sample images (images of different scenes).
One possible way is to bootstrap a video sequence with existing methods
for loopclosure detection.
Such a preprocessed sequence
might not be useful for localization but can provide enough
information to draw positive and negative samples for learning
a whole-image-descriptor with the proposed method.
Several walking, driving, drone videos etc. available on video
sharing websites can be used for learning.
Another way could be to use 3D mesh-models
and render views with nearby virtual-camera locations to obtain positive
samples.
With the advent of crowd sourcing street scenes and availability of
services like mappilary\footnote{https://www.mapillary.com/}, it is easily possible
to assemble a much larger dataset for training. Faster and descriminative learning is even
more crucial when making use of such larger training datasets. We differ this until our future work.

For our experiments be comparable with exisiting methods, we use
the Pittsburgh (Pitts250k) \cite{torii2013visual} dataset which contains 250k
images from Google's street-view engine. It provides multiple
street-level panoramic images for about 5000 unique locations in Pittsburg,
Pennsylvania over several years. Multiple panoramas are available at a particular place ($\approx$ 10m apart) sampled approximately 30 degrees apart along the azimuth.
Another similar dataset is the TokyoTM dataset \cite{torii2013visual}.

\subsection{Training Hyperparameters}
The CNN-learnable parameters are initialized with the Xavier initialization \cite{glorot2010understanding}.
We initialize NetVLAD parameter  $\mathbf{c}_k$ as unit vectors drawn randomly from a surface of a hypersphere.
$b_k$ and $\mathbf{w}_k$ are coupled with $\mathbf{c}_k$ at initialization.
However, as learning progresses these variables are decoupled.
We use the AdaDelta solver \cite{zeiler2012adadelta} with
batch size of $b=4$  (each batch with $m=6$ positive
samples and $n=6$ negative samples)). This configuration
takes about 9GB of GPU memory during training.
We stop the training at 1200 epochs. Our 1 epoch is
500 randomly drawn tuples from the entire dataset.
The learning rate is reduced by a factor of 0.7 if loss
function does not decrease in 50 epochs and a regularization constant is set to
0.001 (to make regularization loss about 1\% of fitting loss).
Data augmentation (rotation, affine
scale, random cropping, random intensity variation) is used for robust learning,
which we begin after 400 epochs. This explains the
rise in the loss function values in our experiments in
Fig. \ref{fig:triplet-vs-allpair-on-pw13-k16},
\ref{fig:triplet-vs-allpair-on-vggblock4-k16},
\ref{fig:triplet-vs-allpair-on-pw7-squash-chnls-k16}.

\begin{figure}
\includegraphics[width=\columnwidth]{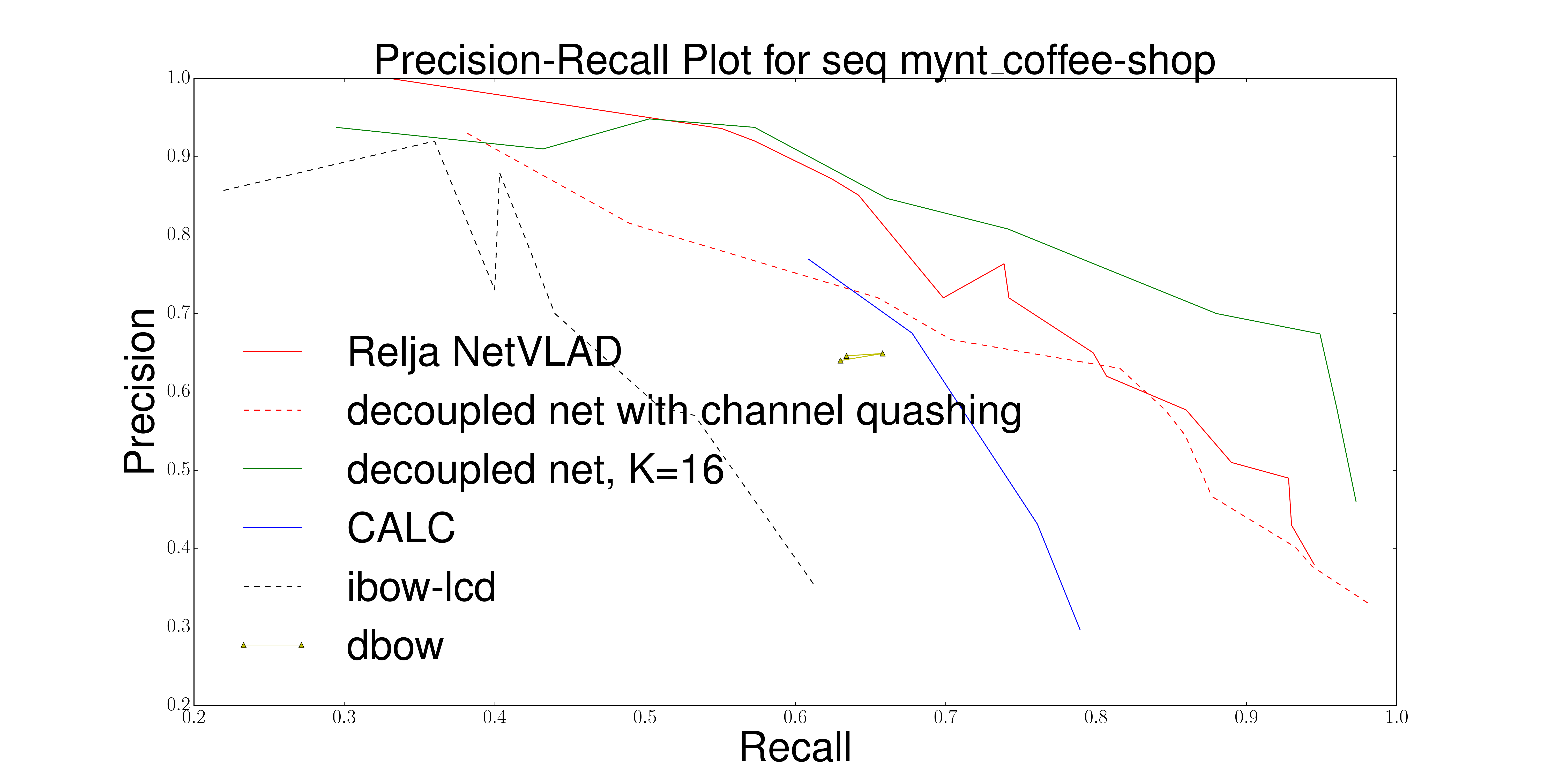}
\caption{Precision-recall plot for the sequence `mynt\_coffee-shop` when compared to manual annotations of loop candidates and
threshold varied. We compare the following methods:
Relja NetVLAD \cite{arandjelovic2016netvlad}, decoupled net with channel squashing (proposed), decoupled net without channel squashing,
CALC \cite{merrill2018lightweight}, ibow-lcd \cite{garcia2018ibow} and DBOW \cite{galvez2012dbow}. }
\label{fig:pr-plot-mynt-coffee-shop-seq}
\end{figure}

\begin{figure}
\includegraphics[width=\columnwidth]{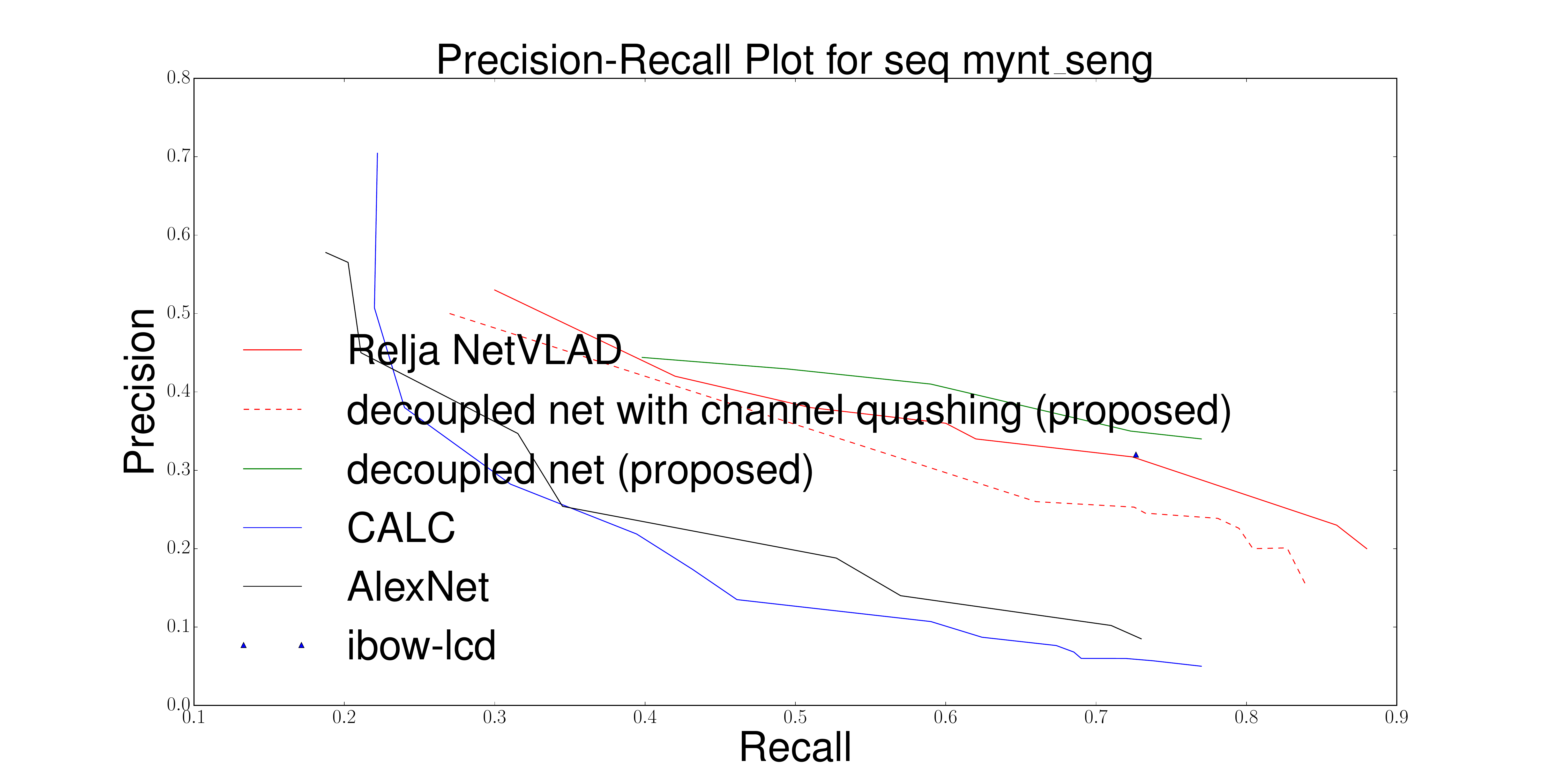}
\caption{Similar to Fig. \ref{fig:pr-plot-mynt-coffee-shop-seq} but for sequence `mynt\_seng`.  }
\label{fig:pr-plot-mynt-seng-seq}
\end{figure}

The output descriptor size is $K \times D$. K is the number of clusters in
NetVLAD, we use K=16,32,64. D is the number of channels of the output CNN.
For both VGG16 and the decoupled network it is 512. Some other
authors notably Arandjejovic \cite{arandjelovic2016netvlad} and Sunderhauf \cite{sunderhauf2015performance} have used
whittening-PCA and gaussian random projections respectively to
reduce the dimensions of the image descriptor from about 64K to 4K.
We suggest to use
learnable squashing channels (to say 32) with channelwise convolutions
before feeding the
pixelwise descriptors to the NetVLAD layer rather than reduce the
dimensionality of the image descriptor. We have
experimentally compared the effect of this channel squashing in
Fig. \ref{fig:pr-plot-mynt-coffee-shop-seq} and
Fig.\ref{fig:pr-plot-mynt-seng-seq}.

\section{Deployment}
\label{sec:deploy}
Our system, which is able to correct drifts of VIOs, recover from long kidnaps
and system failures online in real-time, is available as a add-on to the popular VINS-Fusion.
In general, our system can work with any VIO system.
It is worth nothing that the \textit{maplab} system's \cite{schneider2018maplab} \textit{ROVIOLI}
cannot identify kidnaps and recover from them online, although it can merge multiple
sequences offline (when all the sequences are known).

\begin{figure}
    \centering
    \includegraphics[width=0.9\columnwidth]{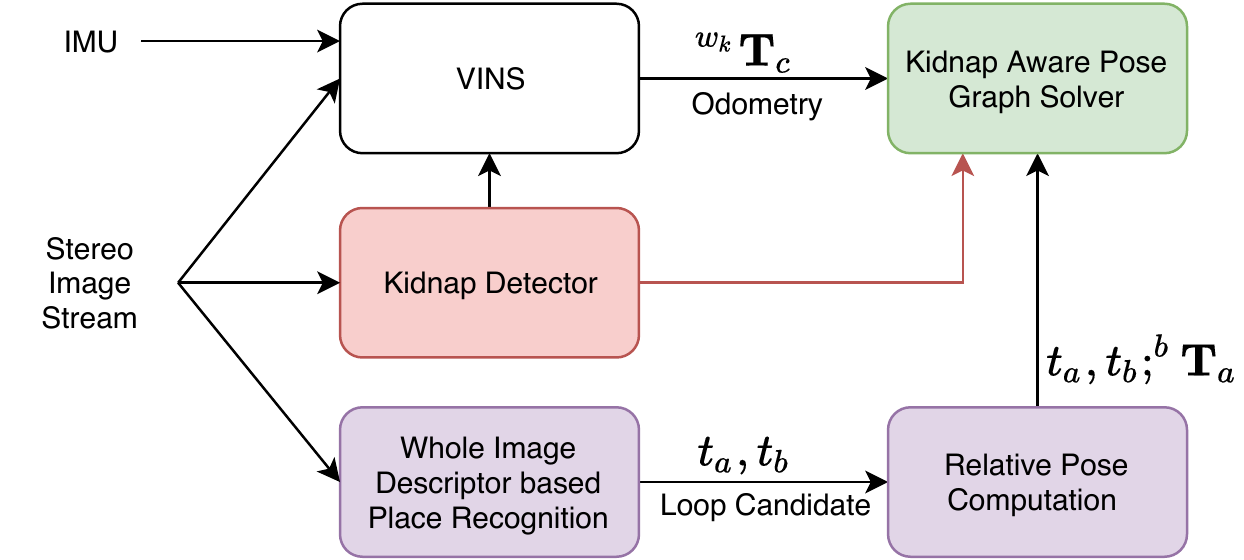}
    \caption{System Overview}
    \label{fig:deploy-full-system}
\end{figure}

In this section we describe our multi-threaded software architecture (Fig. \ref{fig:deploy-full-system}).
We use the consumer-producer paradigm for real-time deployment of our system.
We start by describing the image level descriptor extraction and comparison. After that,
we describe the details of pose computation at the loopcandidate image pair.
Next we describe how our system is able to recover from kidnaps by keeping
track of the disjoint co-ordinate systems and switching off the visual-interial odometry
when no features can be reliably tracked.

\subsection{Descriptor Extraction and Comparison}
\label{sec:desc_extraction_comparison}
We propose a pluggable system to the popular VINS-Fusion\footnote{https://github.com/HKUST-Aerial-Robotics/VINS-Fusion}
by Qin \textit{et al.}\cite{vins-mono}.
Our system receives keyframes from the visual-inertial odometry sub-system
to produce loopclosure candidates.
We use a naive store-and-compare strategy to find loopcandidates.
The descriptors at all previous keyframes, ie. $\eta^{(I_{t})} \ t=1,\ldots,t$ are stored
indexed with time. When a new keyframe
arrives, say $\eta^{(I_{t+1})}$, we perform $\langle \eta^{(I_{t_i})}, \eta^{(I_{t+1})} \rangle \ i=1,\ldots,t-T$.
$T$ is typically 150, ie. ignore latest 150 frames (or 15 seconds) for loopclosure candidates.
These are a measure of likelihoods for loopclosure at each of the keyframe timestamps.
We accept the loopclosure hypothesis if the query score is above a set threshold (fixed
for all the sequences) and if three consecutive queries retrieve
descriptors within six keyframes of the first of the three queries.
In a real implementation, the threshold can be set a little lower
and wrong hypothesis can be eliminated with geometric verifications heuristics.
We note that loopcandidates with large viewpoint difference provide
a formidable challenge for tracked feature matching between the two views.
In this work for comparison of descriptor's precision-recall,
we do not perform any geometric verification.

Our naive comparison (matrix-vector multiplication) takes
about 50 ms for comparison with 4000 keyframes (about 10 min sequence) on a desktop CPU
with image descriptor dimension of 8192 and about 10ms for 512
dimensional image descriptor.
While the comparison times grow unbounded as number of keyframes increase,
the objective of this paper is to demonstrate the representation power of learned
whole image descriptors over
the traditional BOVW on sparse feature descriptor loopclosure detection framework
and the recently proposed CNN-based image descriptors in terms of detection under large viewpoint difference.
Dealing with scalability could be a future research direction. In our
opinion scalability can be achieved by sophisticated product quantization
approaches similar to Johnson \textit{et al.}\cite{faiss2017} or by
maintaining a marginalized set of scene descriptors, along with
scene object labels and dot product comparison on this smaller subset.

\subsection{Feature Matching and Pose Computation}
In this section we describe the task of metric scale relative pose computation given a loopcandidate
pair.
For computation of reliable pose estimates we make use of the GMS-Matcher \cite{bian2017gms}
as a robust correspondence engine. The GMS-matcher propose a simple grid based voting
scheme for finding fast correspondences. This matcher provides reliable matches even under
large viewpoint difference. This can be attributed to the grid-based voting scheme that it implements,
has the effect of eliminating point correspondences if nearby points in one view
do not go to nearby points in the second view. Computationally it takes about
150-200 ms per image pair (image size 640x480).
Since we only need to compute these correspondence on loopcandidates and also
since we use a multi-threaded implementation, this higher computational
time does not stall our computational pipeline.
Note that in the event of loop detections under large
viewpoint difference, the tracked features in a SLAM system do not provide enough
information for feature computation.
Some other related recent works
on CNN based feature correspondence
include InLoc \cite{taira2018inloc}, Neighbourhood consensus network\cite{rocco2018neighbourhood}.
Although these approaches present impressive results, they involve
storage of allpair dense pixel-level features which currently cannot be accomplished in realtime.

Once the feature correspondences are produced we make use of
direct perspective-n-point (PNP) method by Hesch and Roumeliotis \cite{hesch2011direct}
for pose computation \footnote{Implementation from \textit{theia-sfm} (\url{http://www.theia-sfm.org/})}.
We reject a loopcandidate if insufficient number of
correspondences are produces (we use a threshold of 200 correspondences).
We use random sampling and consensus (RANSAC) to make the PNP
robust under spurious feature correspondences that may
occasionally occur. For about 10 RANSAC iteration it takes on  average 5 ms
to compute the pose from approximately 2000 feature correspondences.
We make use of stereo geometry for computing 3D points at loopclosures.

In cases where the queue size is small and computational resources sufficient
we also compute the pose by iterative closest point (ICP) method in addition to the PNP.
The 3D points for both the images of the loopcandidates is obtained from stereo geometry in
this case.
If the pose computations are not consistent from the two methods we reject the loopcandidate.

\subsection{Kidnap Detection and Recovery}
In our system we also deal with a kidnap recovery mechanism. By kidnap we refer to the
camera's view being blocked and the camera teleported to another location several 10s to 100s
of meters away in 10s to 100s of seconds. The teleported location may or may not be a
previously seen location.
We make use of a simple criterion like the current number of tracked features
falling to a very low value (like less than 10) to
determine if the camera system is kidnapped. Once we determine that the camera
system has been kidnapped we stop the visual inertial odometry subsystem.
When sufficient number of features are again being tracked we reinitialize the odometry/sensor fusion
system. It is to be noted that in such a case it starts with a new co-ordinate reference.
Hence forth we refer to the new co-ordinate systems as world-0 ($w_0$), world-1 ($w_1$),
world-2 ($w_2$) and world-k ($w_k$) in general. We denote the nodes $n$ by a superscript to identify
the world that it is in. For instance $n_i^{(k)}$, means the $i^{th}$ node (i is the global index of the node)
is in the $k^{th}$ world. The odometry pose of the node is denoted as $^{(k)}\mathbf{T}_i$.

\begin{figure}
    \centering
    \includegraphics[scale=0.5]{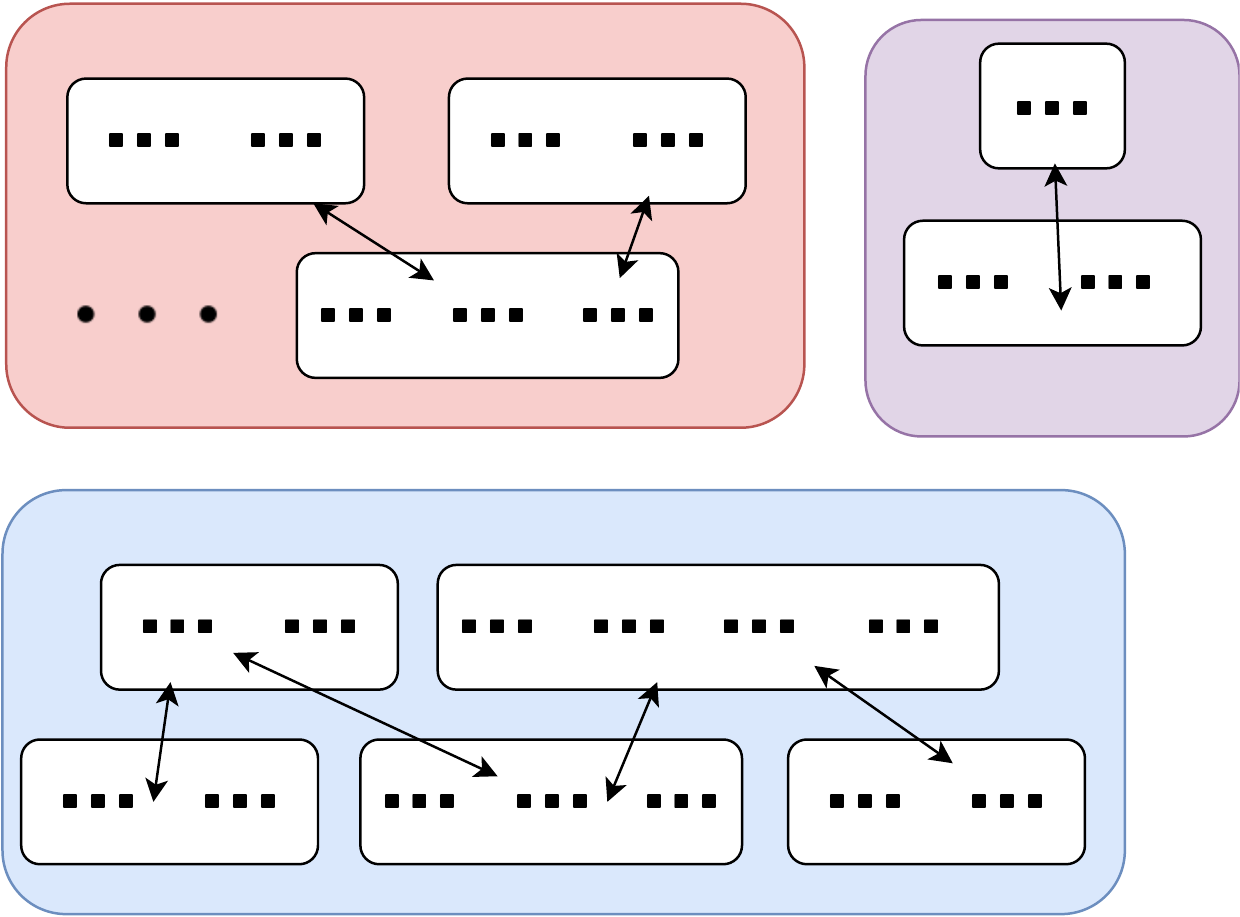}
    \caption{The solid black squares represent the nodes in the pose graph. Arrows
    show the loopcandidates. The white rectangles show each of the individual worlds.
    The colored rectangular enclosures are the worlds belonging to the same set.
    }
    \label{fig:disjoint-set-ds}
\end{figure}

The incoming loopcandidate pair can be categorized into two kinds (refer to figure \ref{fig:disjoint-set-ds}
for a visual explaination).
a) Intra worlds (eg. $n_i^{(k)} \leftrightarrow	 n_j^{(k)}$ )
and b) Inter worlds ($n_i^{(k)} \leftrightarrow	 n_j^{(k')}$).

\subsubsection{Direct Pose Computation Between $w_k$ and $w_{k'}$}
\label{sec:direct-worldpose-infer}
The inter-world loopcandidates $n_i^{(k)} \leftrightarrow	 n_j^{(k')}$
can be used to infer the relative poses between the co-ordinate system $w_k$ and $w_{k'}$.
In this section, we describe the computation of the relative pose between the worlds $k$ and $k'$, ie.
$^{(k)}\mathbf{T}_{(k')}$ from the relative pose between the two nodes ($^i\mathbf{T}_j$) and the
odometry poses of the nodes
in their respective worlds ie. $^{(k)}\mathbf{T}_i$ and $^{(k')}\mathbf{T}_j$ respectively.
\begin{equation}
^{(k)}\mathbf{T}_{(k')} = ^{(k)}\mathbf{T}_i \times ^i\mathbf{T}_j \times ( ^{(k')}\mathbf{T}_j )^{-1}
\end{equation}

\subsubsection{Indirect Pose Computation Between $w_k$ and $w_{k_1}$}
\label{sec:indirect-worldpose-infer}
It is easy to see that if we have two inter-world loopcandidates like:
 $n_i^{(k)} \leftrightarrow	 n_j^{(k')}$ and  $n_{i_1}^{(k_1)} \leftrightarrow	 n_{j_1}^{(k')}$,
 the three worlds $w_k$, $w_{k'}$ and $w_{k_1}$ are said to be in same set. It is also
 possible to indirectly infer the relative pose between worlds $w_k$ and $w_{k1}$
 even though no loopcandidate exists between these two sets. This
 estimate is needed for correctly initializing the poses to solve the pose graph
 optimization problem.

We make use of the data structure disjoint sets \cite{cormen2009introduction} to maintain
the world information that are in the same set. The advantage of the disjoint set datastructure
is it provides for a constant time set-union and sub-linear time set-association query.
Each world starts in its own set,
everytime we encounter the inter-world looppair we merge these two sets of the worlds
into a single set.

When we assert that two different worlds are in the same set, we imply
that a  relative transform between these two worlds can be determined.
However that a loopcandidate between
these two pairs of worlds may or may not exist. In case no loopcandidate exists between
the two worlds but these two worlds are in the same set, the relative poses between the worlds
can be determinted by finding a graph-path between the two worlds and chaining the
relative pose estimates between the adjacent world pairs in the path.

In a general scenario, this can be accomplished by constructing a directed graph of the worlds
with nodes being the worlds in the same set and edges being the relative poses between these two worlds,
ie. $^{(k)}\mathbf{T}_{(k')}$.
A breadth-first search on this graph is sufficient to determine an estimate of relative
poses between arbitary pairs of worlds by chaining the relative poses of the path generated by the graph search.

\subsection{Implementation Details}
Our full system employs multiple threads. It uses the producer-consumer programming
paradigm for managing and processing the data. In our system, thread-1 produces image descriptors
of all incoming keyframe images. Thread-2 consumes the image descriptors to produce candidate
matches. Thread-3 consumes the candidate matches to produce feature correspondences.
Thread-4 uses the feature correspondence to produce the relative pose $^i\mathbf{T}_j$.
Thread-5 monitors the number of tracked features to know if the system has been kidnap.
Thread-6 uses the loopcandidates and their relative poses to construct the disjoint
set data structure and maintain the relative poses between multiple co-ordinate systems
as detailed in Sec. \ref{sec:direct-worldpose-infer}
and \ref{sec:indirect-worldpose-infer}. Thread-7 incrementally constructs and solves
the pose graph optimization problem while carefully initializing the initial poses
and making use of poses between the worlds. Our pose graph solver is based upon
the work of Sunderhauf \textit{et al.} \cite{sunderhauf2012switchable}.
A separate thread is used
for visualizing the poses. Even though we use 7 threads, the effective load factor on the
system is about 2.0. This means about two cores are occupied by our system (this
does not include the processing for VINS-Fusion Odometry System).

We demonstrate the working of our full system (see Fig. \ref{fig:kidnap-screenshot}). It is worth
noting that in addition to reducing the drift on account of loopclosures, our implementation
 can reliably identify and recover from kidnap scenarios
lasting longer than a minute online in realtime.
We attribute such robustness to the high recall rates of the
NetVLAD based image descriptor engine.
The results in regard to the operation of the full system
can be found in the attached video.
We record our own data for demonstating the kidnap cases. Some of the
kidnap cases are labelled hard which include the need for indirect inference which is not
available in the previous slam systems including the relocalization system
from Qin \textit{et al.}\cite{qin2018relocalization}.

\section{Experiments}
\label{sec:all_experiments}
In this section we describe our experiments. We evaluate the
effect of using the allpair loss function
compared to the commonly used triplet loss function (Sec. \ref{sec:exp-effect-of-allpair-loss}), while keeping the same backend CNNs.
Next the effect on running time and memory consumption by the use of
decoupled convolutions are tabulated (Sec. \ref{sec:exp-flops}).
In Sec. \ref{sec:exp-precision-recall-vpr-seq} we evaluate the precision-recall performance
of the proposed algorithm with other competing methods in the SLAM
community and in the visual place recognition community.
Finally in Sec. \ref{sec:exp-online-loop-detections} we evaluate the performance of the proposed
method on a real world SLAM sequences captured
under challenging conditions especially revisits
occuring with non-fronto parallel configurations
and in-plane rotations. For demonstration of our full system on real world sequences
refer to the video attachment.

\subsection{Evaluation Metrics for Loss Function}
\label{sec:exp-effect-of-allpair-loss}
We evaluate the effect of the proposed loss function on NetVLAD compared to the original triplet loss which was used in \cite{arandjelovic2016netvlad}.
We evaluate our proposed loss function using
a) relative loss declines b) number of
correctly identified pairs from the validation tuple.
We also plot the variance dot products of the positive samples amongst themselves.
For validation
we use the Pitts30K dataset and for
training we use the TokyoTM dataset.

Similar to the training tuple, a validation tuple contains a query image, nN ($=6$) negative samples
and nP ($=6$) positive samples.
For evaluation, we propose to
count the number of correctly
identified image pairs
which were actually similar  to query images (ground truth)
and identified as similar based on the image descriptor dot product
scores. Ideally, the evaluation metric
should be the SLAM sequences however it becomes infeasible to
evaluate with SLAM sequence every say 20 epochs so we stick
to this workaround.
We plot these metrics for the training data and
the validation data as the iterations
progress. See Fig. \ref{fig:triplet-vs-allpair-on-pw13-k16},
\ref{fig:triplet-vs-allpair-on-vggblock4-k16},
\ref{fig:triplet-vs-allpair-on-pw7-squash-chnls-k16} for
the plots.
The summary of the observation:
\begin{itemize}
\item Using the same backend convolutional network, the same parameters of the netvlad layer, and
same learning hyperparameters, the network trained with the proposed
all-pair loss function performs better as evaluated against the validation metric, count of pairs
correctly identified.
\item It can also be inferred that the gradients obtained from the use of
proposed all-pair loss function are more stable, hence the faster convergence.
\item Our all pair loss function was found to perform better even when using the
decoupled convolutions, decoupled convolution with channel quashing vs the
original VGG16 network.
\end{itemize}

\subsubsection{Number of Zero Loss Tuples}
\begin{figure}
   \centering
   \includegraphics[scale=0.30]{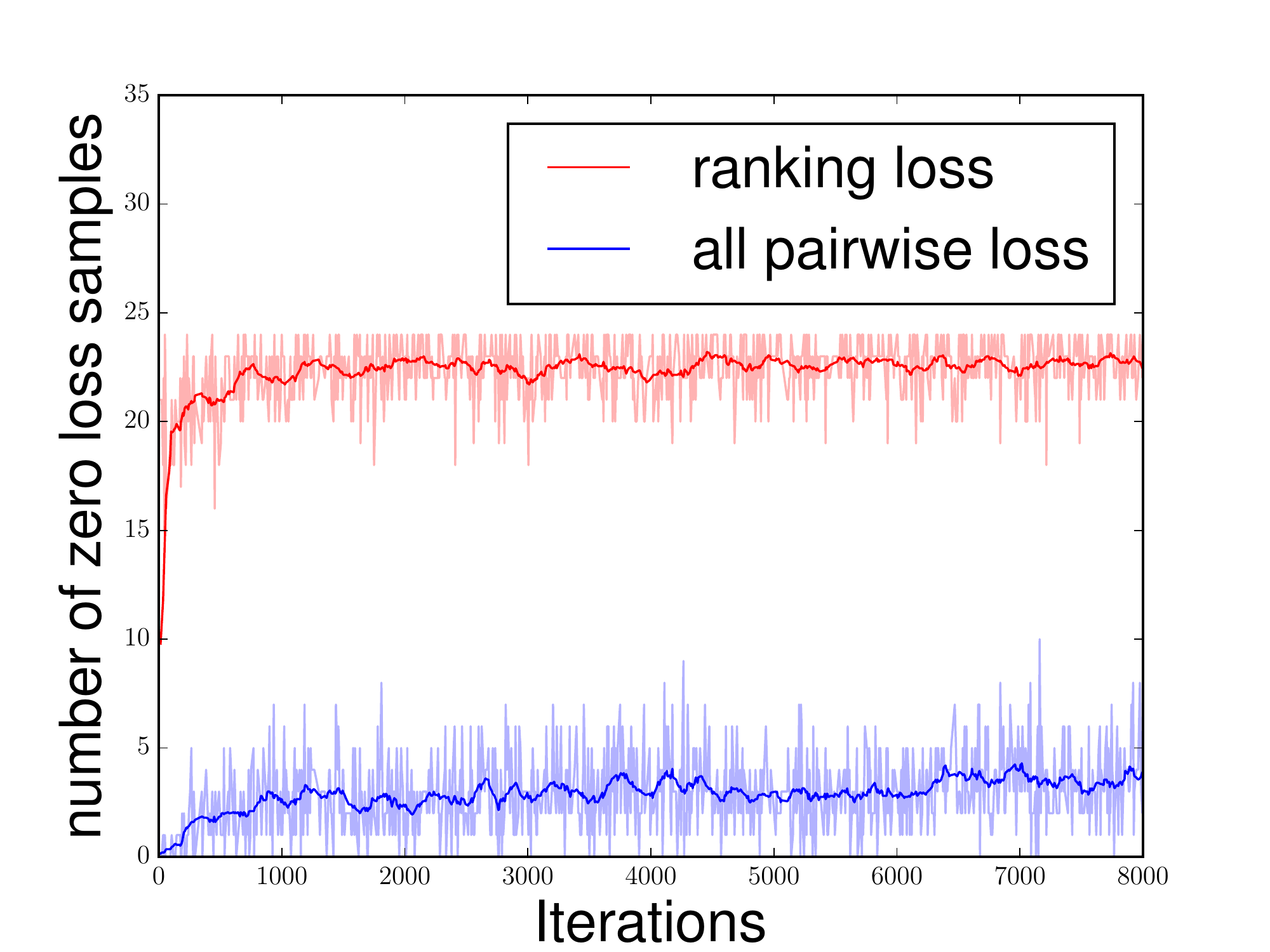}
 \caption{The number of batches with zero loss as iterations progress   for  learning with proposed cost function (in blue, Eq.  \ref{eq:cost_final}) compared    to using the triplet ranking loss \cite{arandjelovic2016netvlad}   (in red, Eq. \ref{eq:netvlad_original}).This experiments
   used a batch size of 24 with gradient cummulation. Having a higher
   count for zero-loss samples is detrimental to learning as it leads to    zero-valued gradients.
   Best viewed in color.}
   \label{fig:batch-zero-loss-compare}
\end{figure}

In this experiment, we train with batch size 24. But since this
won't fit in the GPU memory we use gradient cummulation.
We plot the number of zero loss sample as iterations progress in
Fig. \ref{fig:batch-zero-loss-compare}. When using the
triplet loss, we get more number of zero loss samples.
This results in zero-gradient updates and hence slow learning
compared to proposed allpairloss.
This can be attributed to allpairloss function being harder to satisfy resulting in
better gradients during training.

\subsubsection{Spreads of Positive and Negative Samples}
\begin{figure}
   \centering
   \includegraphics[scale=0.30]{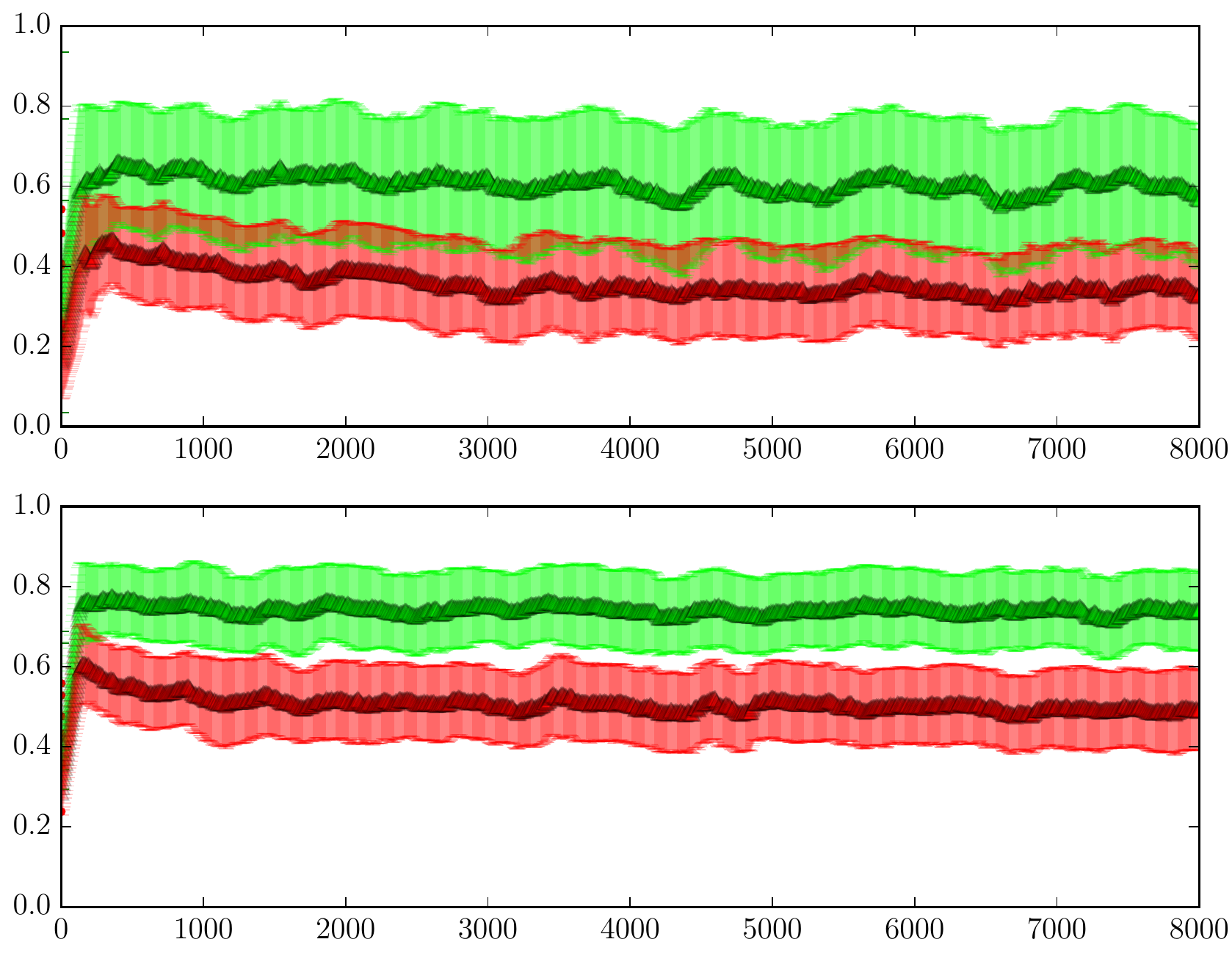}
   \caption{Showing spreads ($\mu \pm \sigma$)  of $\langle \eta_q, \eta_{P_i} \rangle$ (in green) and spreads of $\langle \eta_q, \eta_{N_i} \rangle$ (in red) as the learning progresses. Fig. \ref{fig:pos-set-dev-analysis} (top) corresponds to  \cite{arandjelovic2016netvlad}, with the triplet loss function.
Fig. \ref{fig:pos-set-dev-analysis} (bottom) corresponds to the proposed allpair loss function. We observe a lower spread amongst positive samples and  larger separation between positive and negative samples.}
 \label{fig:pos-set-dev-analysis}
\end{figure}

As experimentally observed in Fig. \ref{fig:pos-set-dev-analysis},
the use of proposed allpair loss function results in a
more discriminative image descriptor as compared
to the network trained with the triplet loss.
We observe a lower spread amongst the positive samples and a larger
separation in positive and negative samples. This has the effect that the
deployment as loopclosure module being less sensitive to slight changes
in dot product thresholds.

\subsubsection{TripletLoss vs AllpairLoss on Decoupled Net}
\begin{figure*}
\centering
\begin{subfigure}
	\centering
	\includegraphics[width=0.4\textwidth]{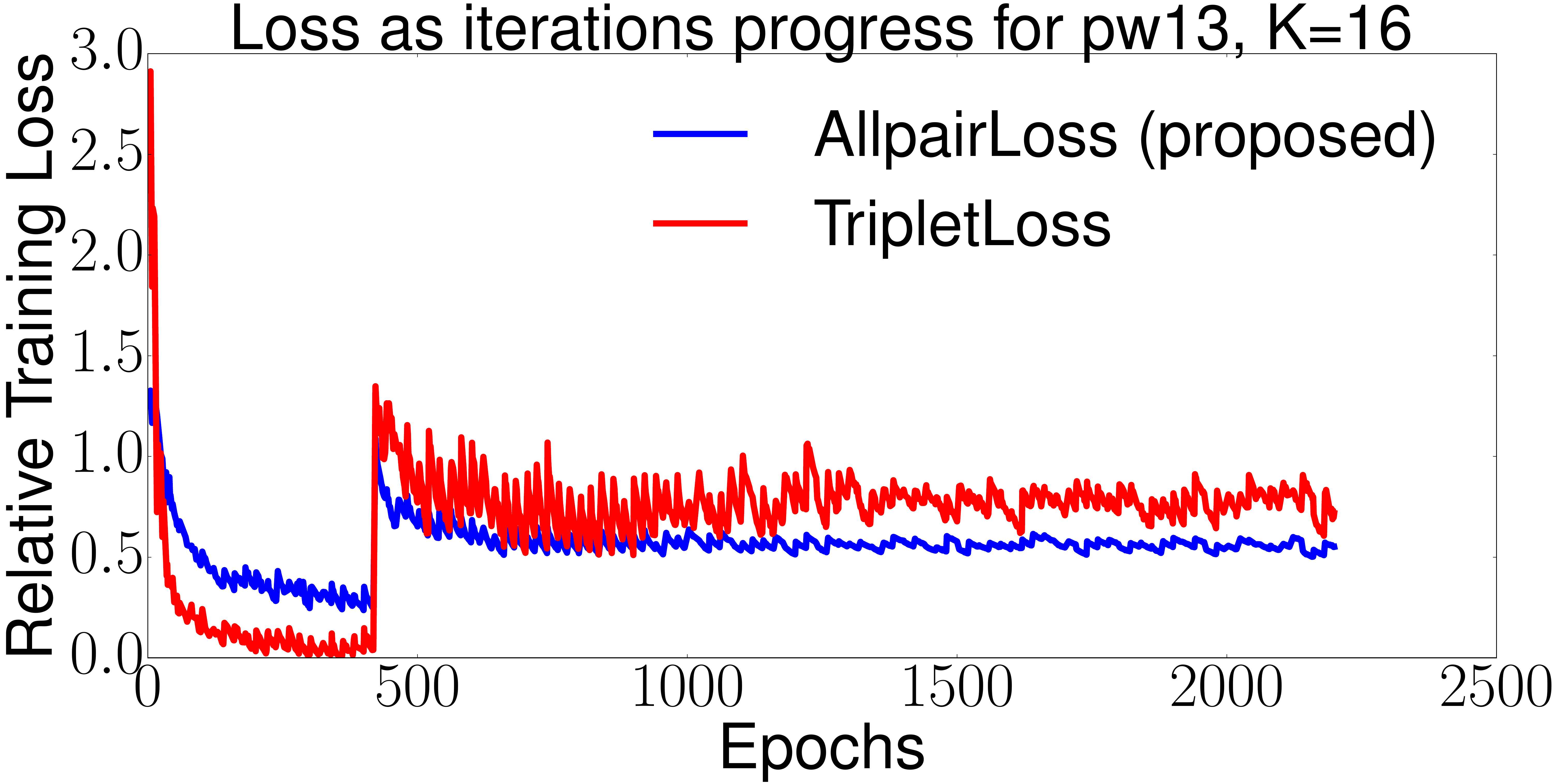}
\end{subfigure}
\begin{subfigure}
	\centering
	\includegraphics[width=0.4\textwidth]{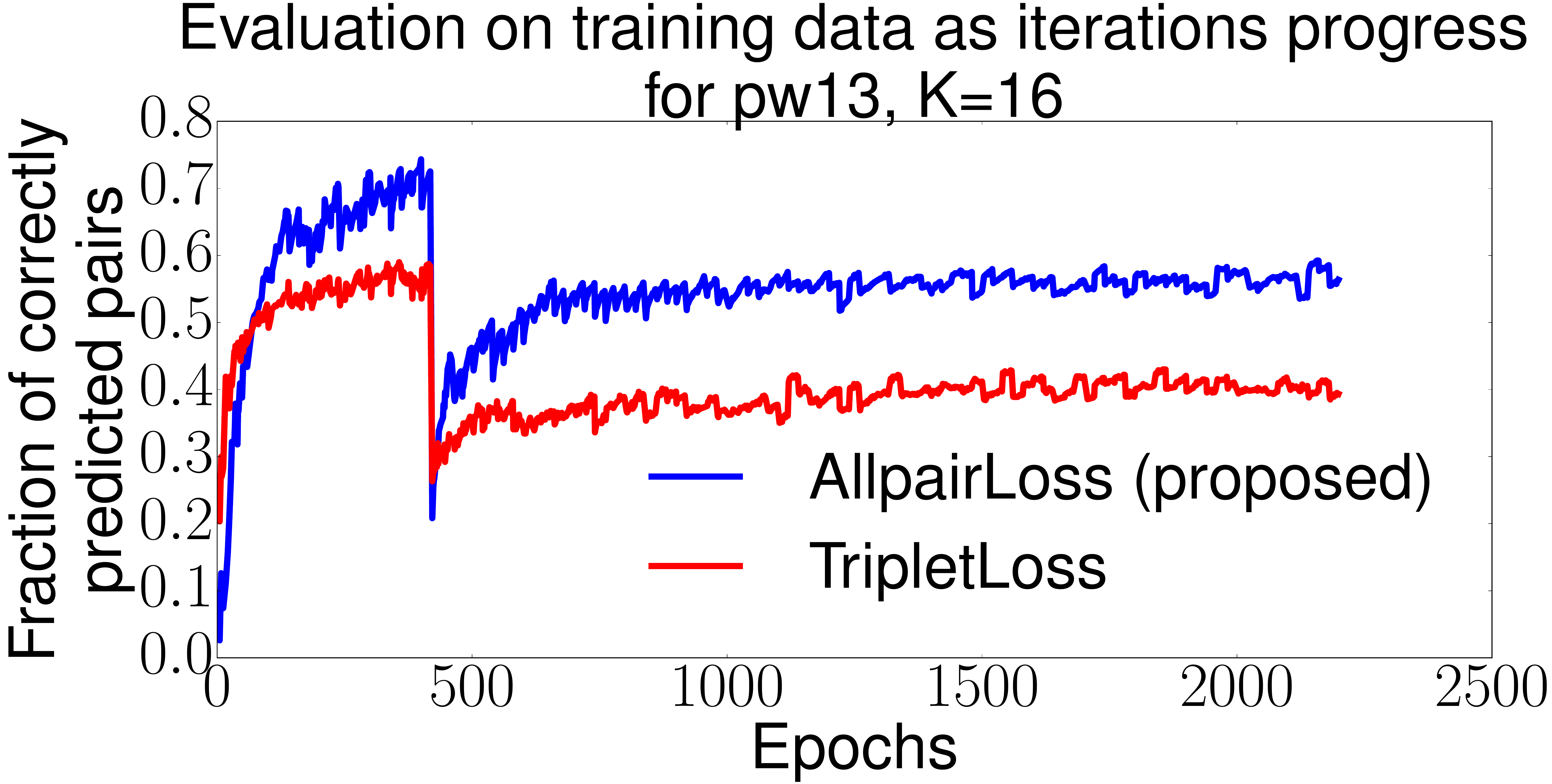}
\end{subfigure}
\begin{subfigure}
	\centering
	\includegraphics[width=0.4\textwidth]{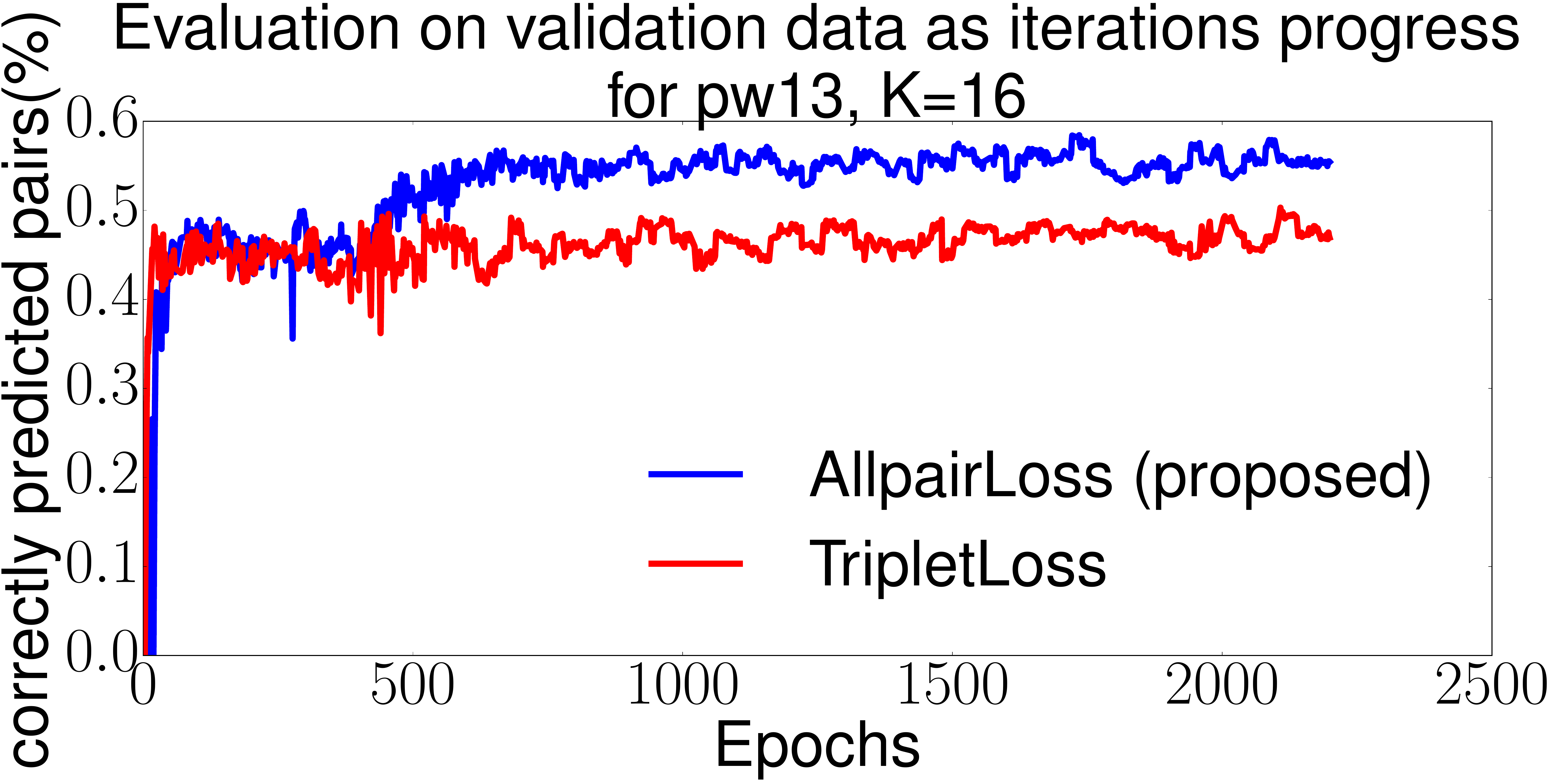}
\end{subfigure}
\begin{subfigure}
	\centering
	\includegraphics[width=0.4\textwidth]{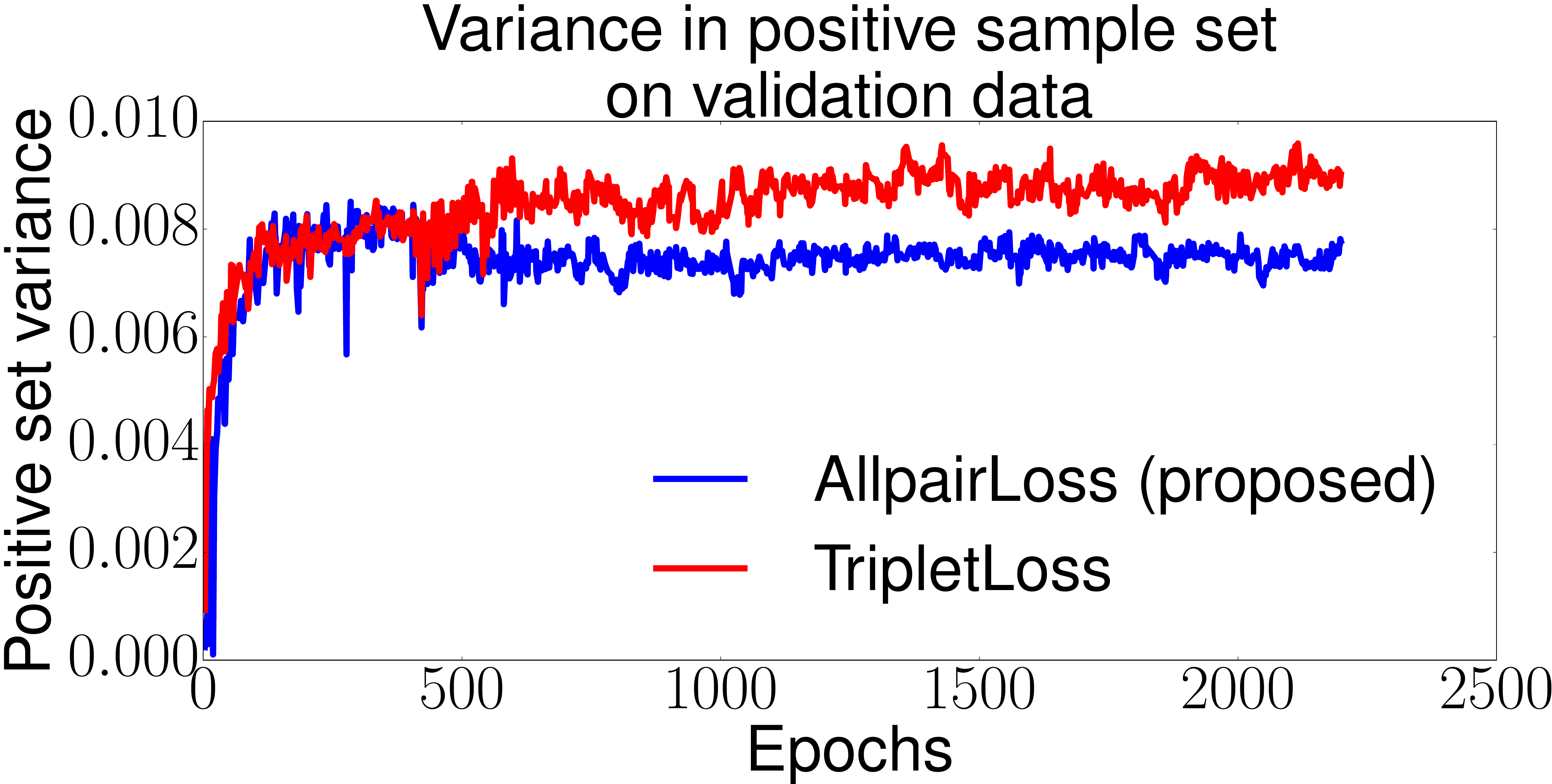}
\end{subfigure}
\caption{Comparing the effect of using allpairloss and tripletloss for
training with decoupled net (deepest layer) with K=16. (a) Shows
the relative training loss as iterations progress (lower is better). We show the evaluation metric, ie. the
count of correctly identified pairs in (b) and (c) for
training data and a separate validation data (higher is better). (d) show
the variance in the positive set in dot product space
as iterations progress (lower is better). }

\label{fig:triplet-vs-allpair-on-pw13-k16}
\end{figure*}

We compare the effect of different loss function when using
the decoupled network. The network trained with
allpair loss is able to correctly identify almost 60\%
of the pairs from the tuples drawn from the validation data,
compared to when network was trained with tripletloss
which is able to identify about 35\% correctly under identical
conditions.
See Fig. \ref{fig:triplet-vs-allpair-on-pw13-k16}.

\subsubsection{TripletLoss vs AllpairLoss on VGG16 Net}
\begin{figure*}
\centering
\begin{subfigure}
	\centering
	\includegraphics[width=0.4\textwidth]{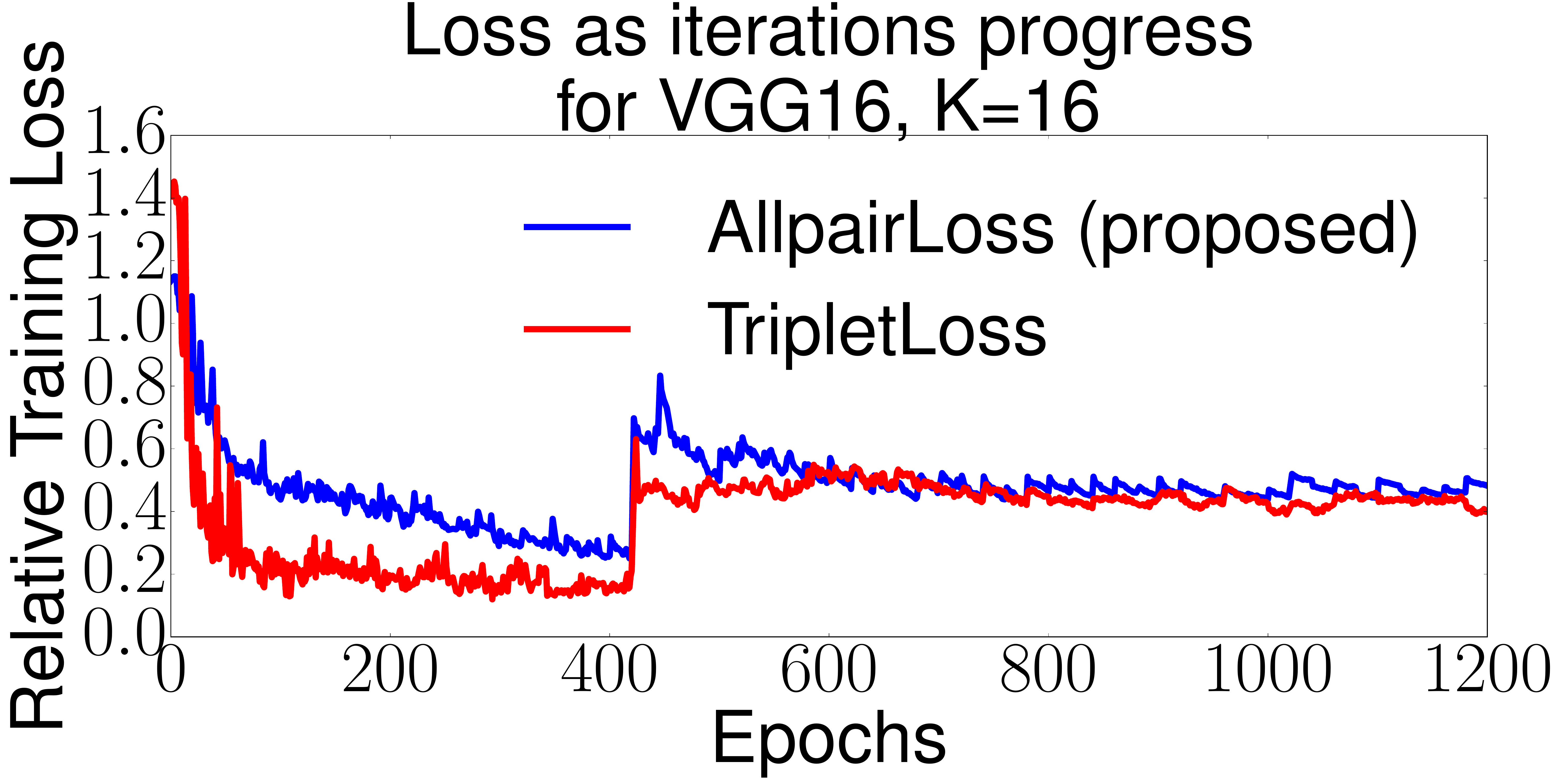}
\end{subfigure}
\begin{subfigure}
	\centering
	\includegraphics[width=0.4\textwidth]{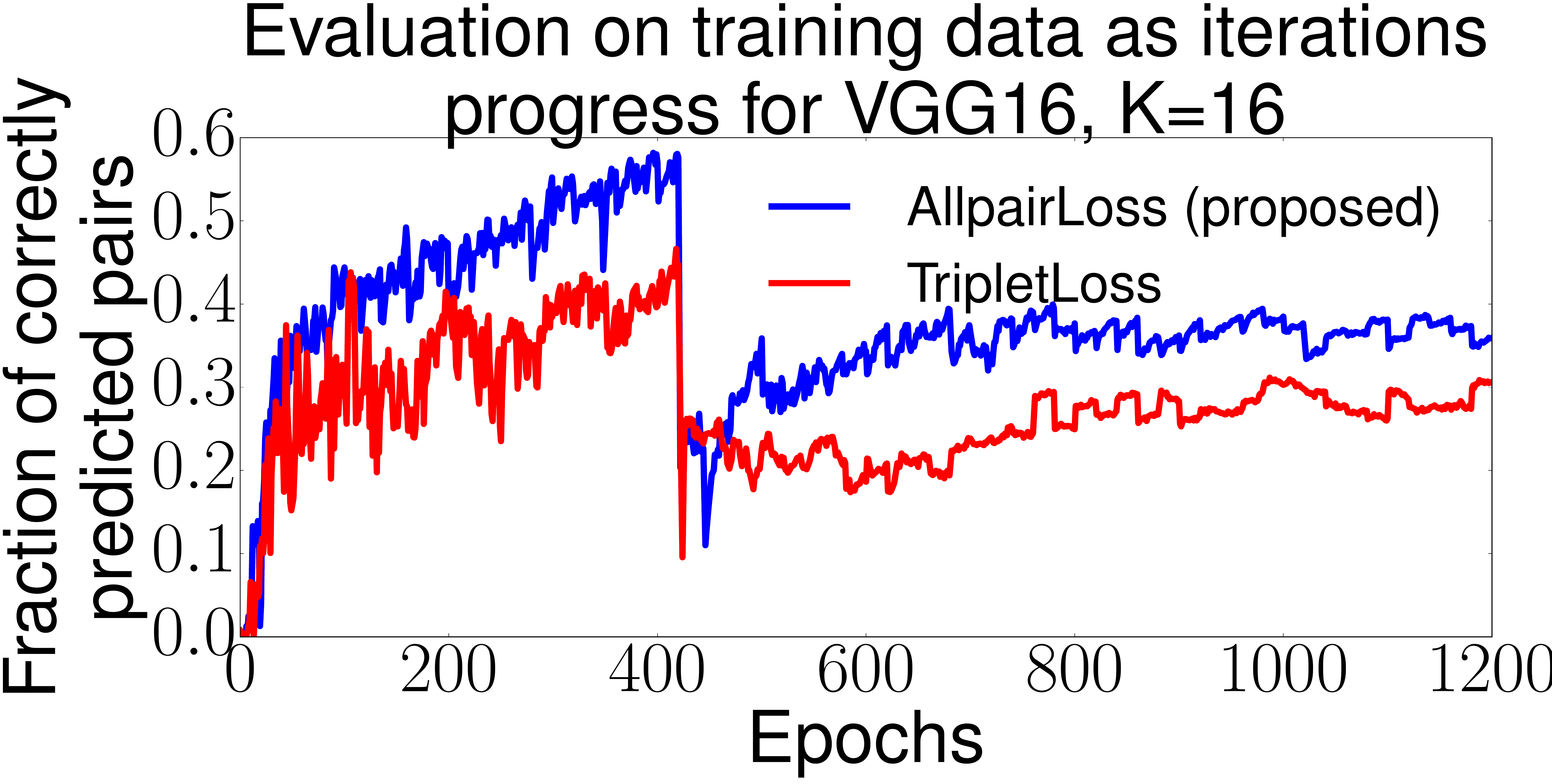}
\end{subfigure}
\begin{subfigure}
	\centering
	\includegraphics[width=0.4\textwidth]{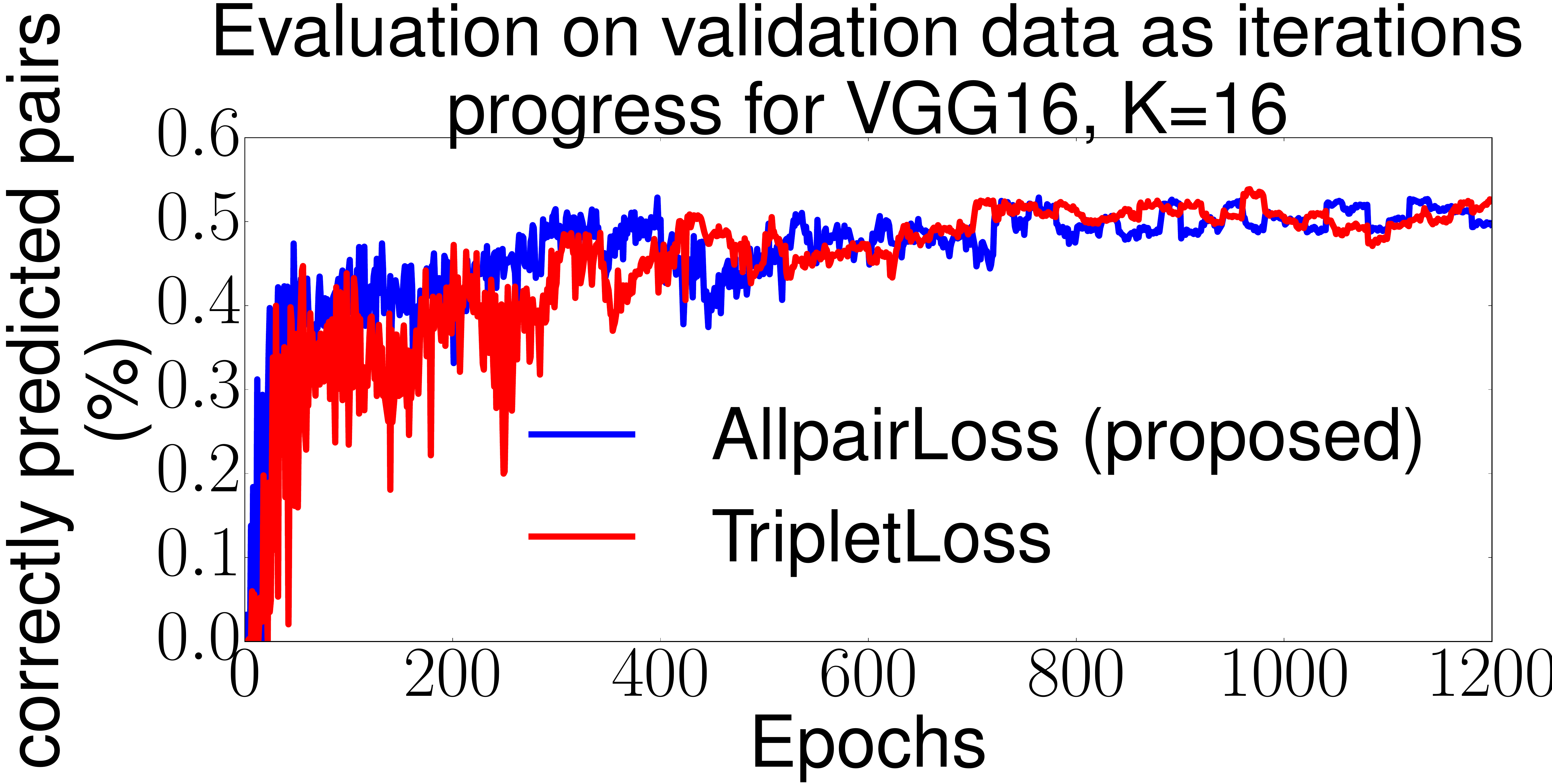}
\end{subfigure}
\begin{subfigure}
	\centering
	\includegraphics[width=0.4\textwidth]{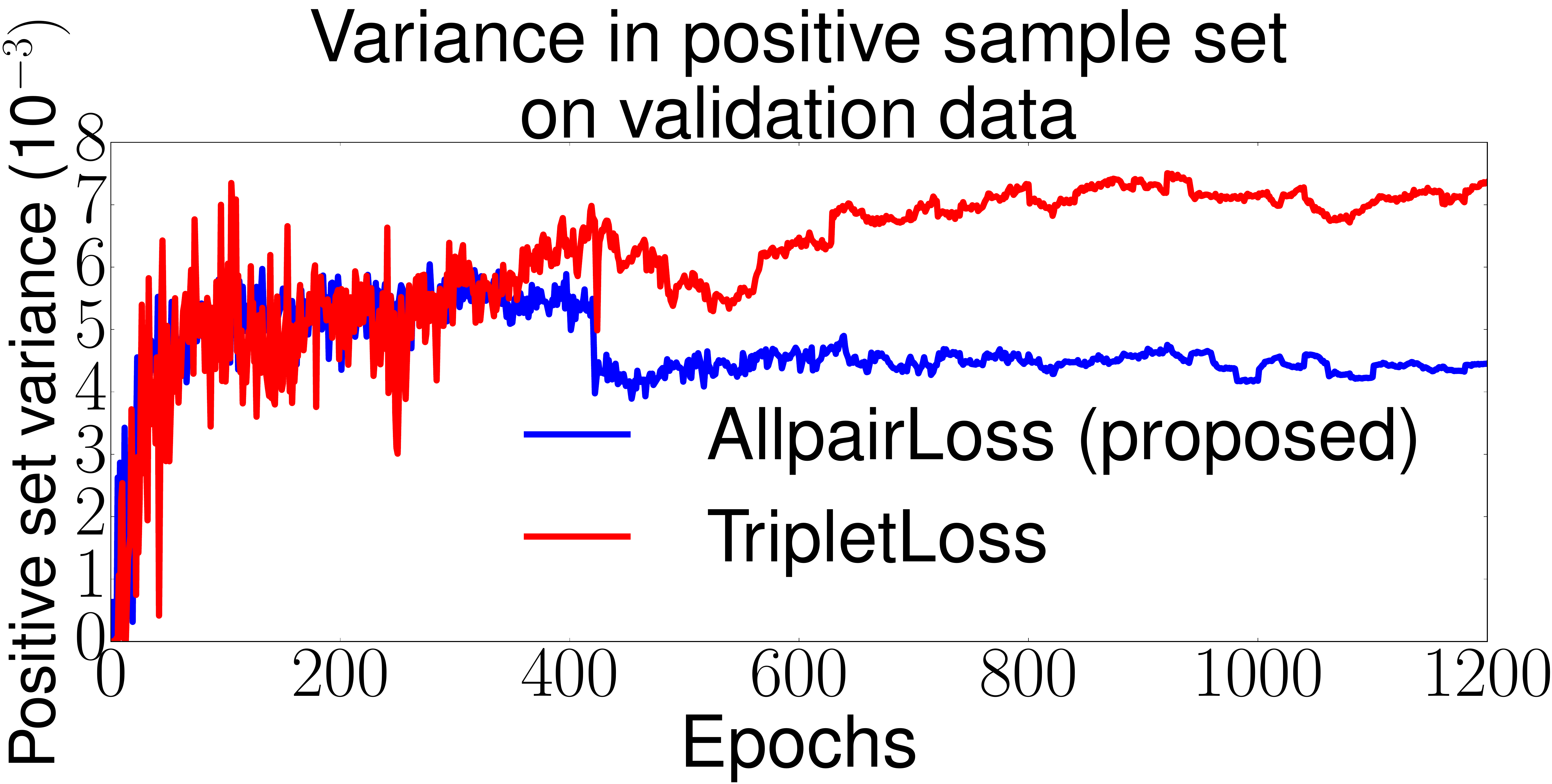}
\end{subfigure}
\caption{Effect of tripletloss and allpairloss with backend CNN
as the VGG16 net with K=16. (a) shows the relative training loss as iterations progress (lower is better). (b) and (c)
shows the training and validation evaluation metric (higher is better).
Evaluation metric is the percentage of pairs correctly identified.
(d) shows the variance of positive set descriptors in dot product space.}
\label{fig:triplet-vs-allpair-on-vggblock4-k16}
\end{figure*}

Even when trained with VGG16 as the backend CNN, allpairloss
performed better than the tripletloss under identical training
conditions.
See Fig. \ref{fig:triplet-vs-allpair-on-vggblock4-k16}.

\subsubsection{TripletLoss vs AllpairLoss on Decoupled Net with Channel Squashing}
\begin{figure*}
\centering
\begin{subfigure}
	\centering
	\includegraphics[width=0.3\textwidth]{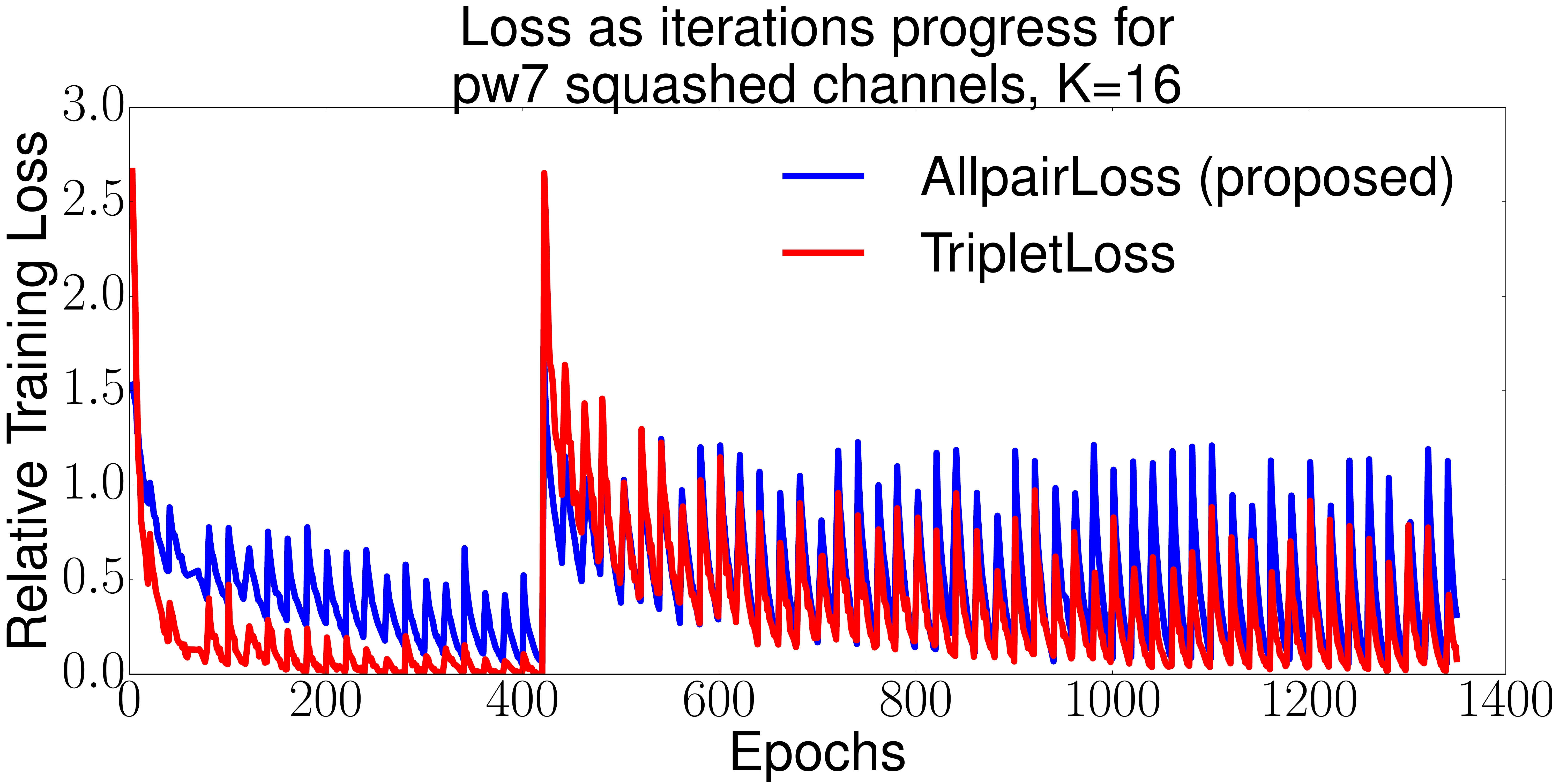}
\end{subfigure}
\begin{subfigure}
	\centering
	\includegraphics[width=0.3\textwidth]{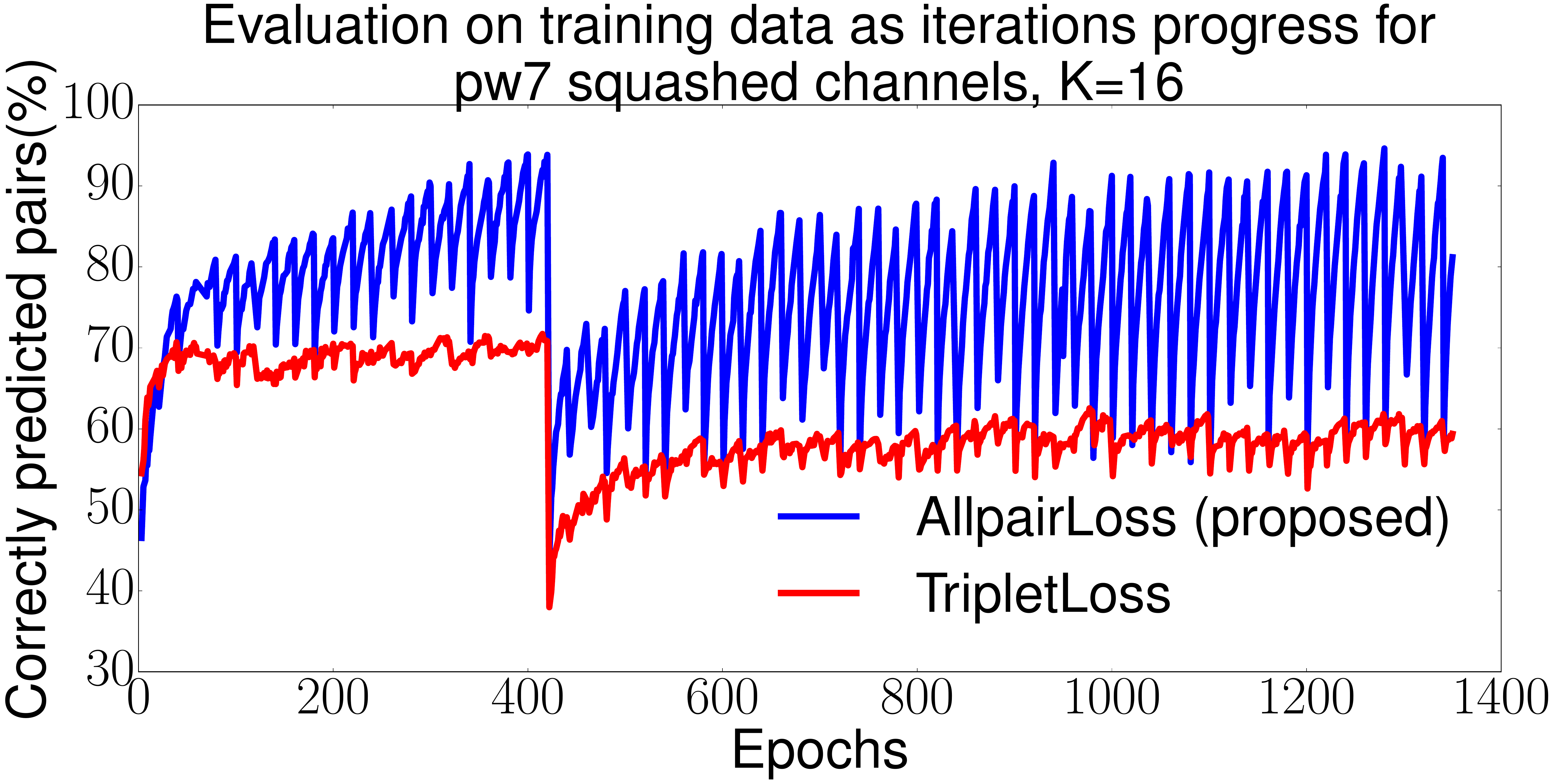}
\end{subfigure}
\begin{subfigure}
	\centering
	\includegraphics[width=0.3\textwidth]{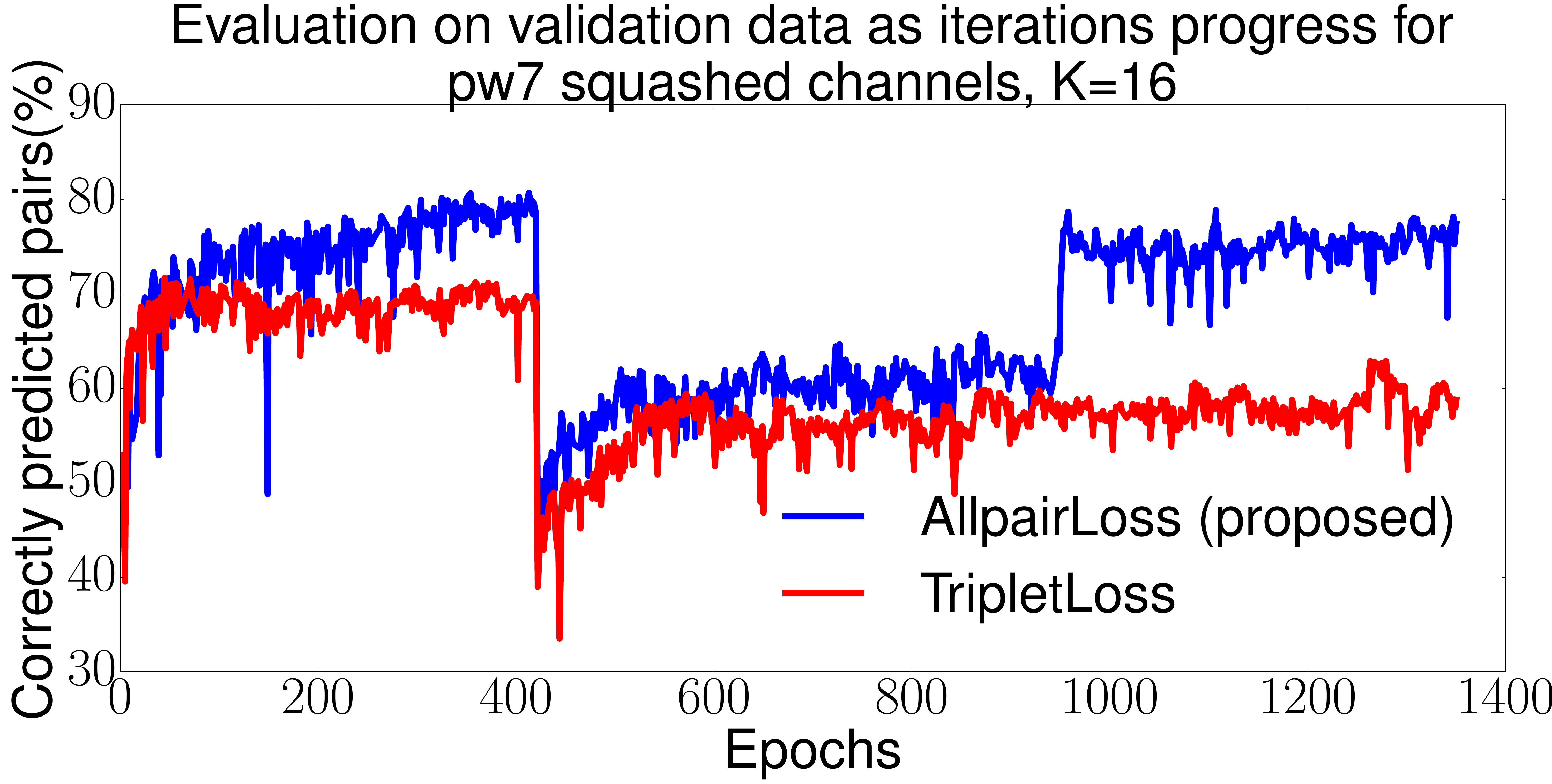}
\end{subfigure}
\caption{ Effect of tripletloss vs allpairloss for decoupled net, K=16
with channel squashing. The descriptor size in this case was just 512.
Arguably the learning in this case can be improved with
lower learning data due to the osscillating losses
we observe. (a) shows the relative training loss. (b) and (c) shows the
percentage of pairs correctly identified for training
and a separate validation data. }
\label{fig:triplet-vs-allpair-on-pw7-squash-chnls-k16}
\end{figure*}

When using decoupled network with channels squashing (for dimensionality reduction) we observe a better performance
when trained with allpairloss. In this configuration the
descriptor size is just 512 per image. The training was
arguably more unstable in this case (we observe oscillations).
Possibly with a lower learning rate this effect can be reduced.
See Fig. \ref{fig:triplet-vs-allpair-on-pw7-squash-chnls-k16}.

\begin{figure*}
\centering
\begin{subfigure}
	\centering
	\includegraphics[width=0.3\textwidth]{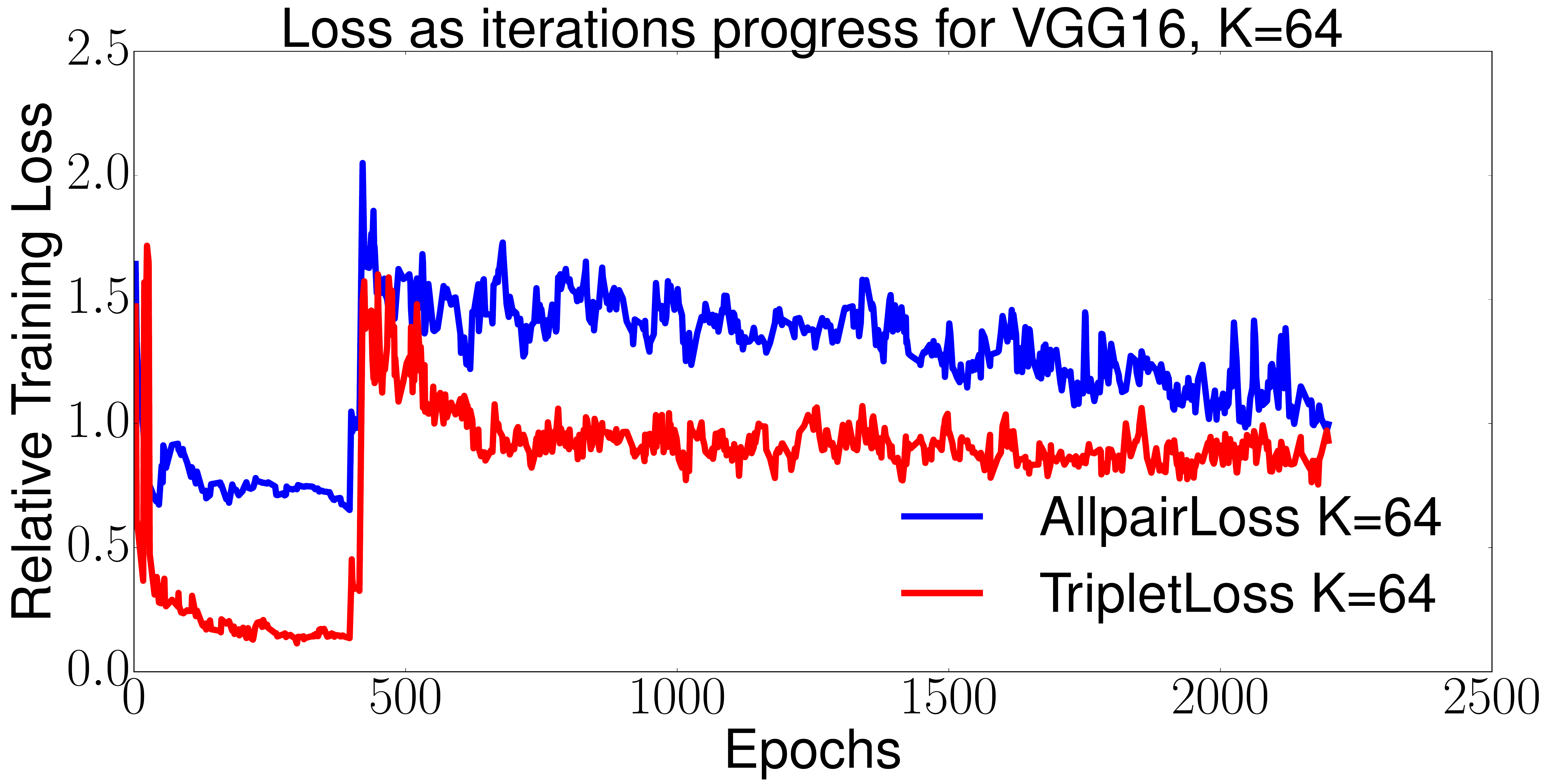}
\end{subfigure}
\begin{subfigure}
	\centering
	\includegraphics[width=0.3\textwidth]{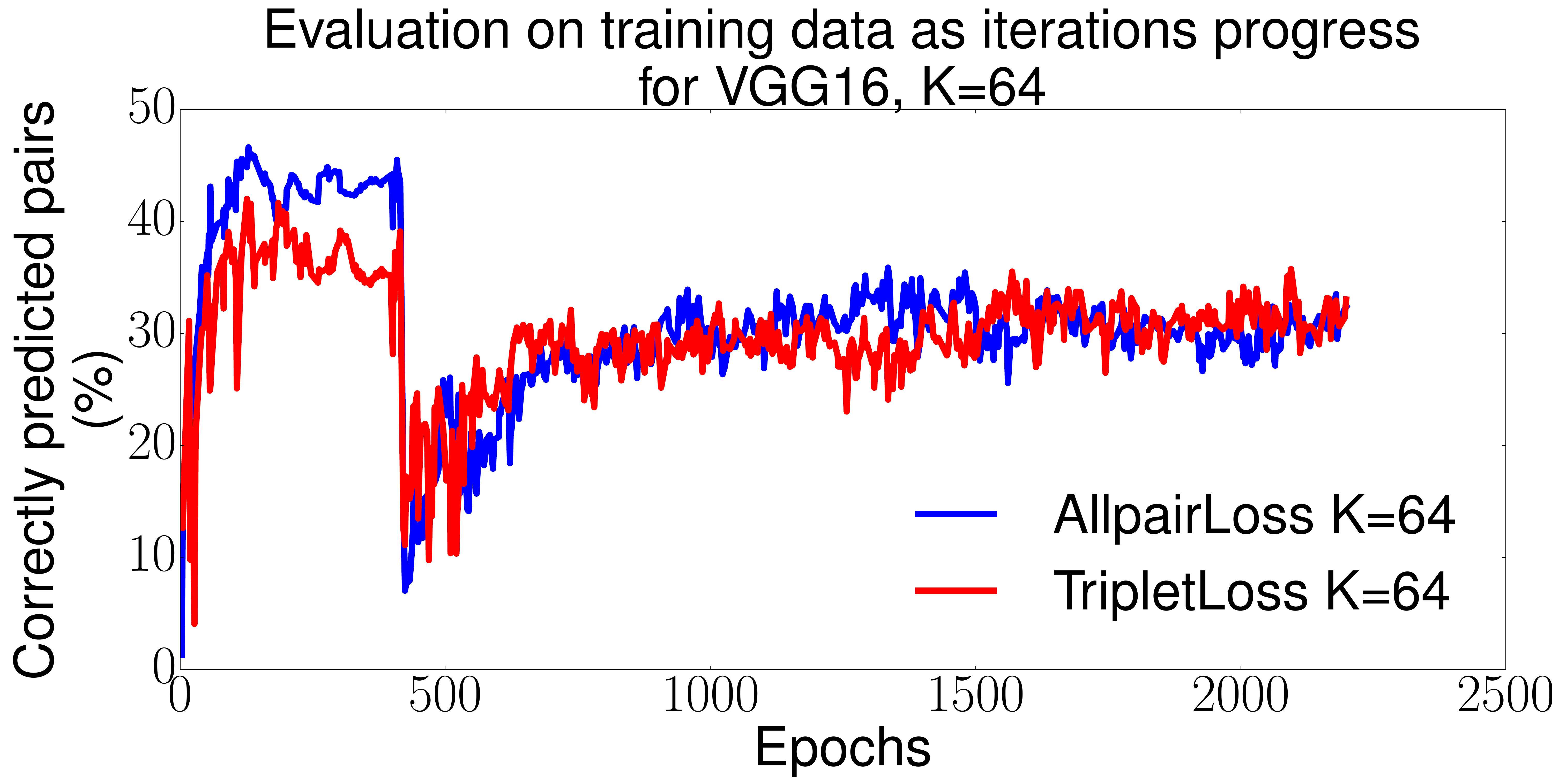}
\end{subfigure}
\begin{subfigure}
	\centering
	\includegraphics[width=0.3\textwidth]{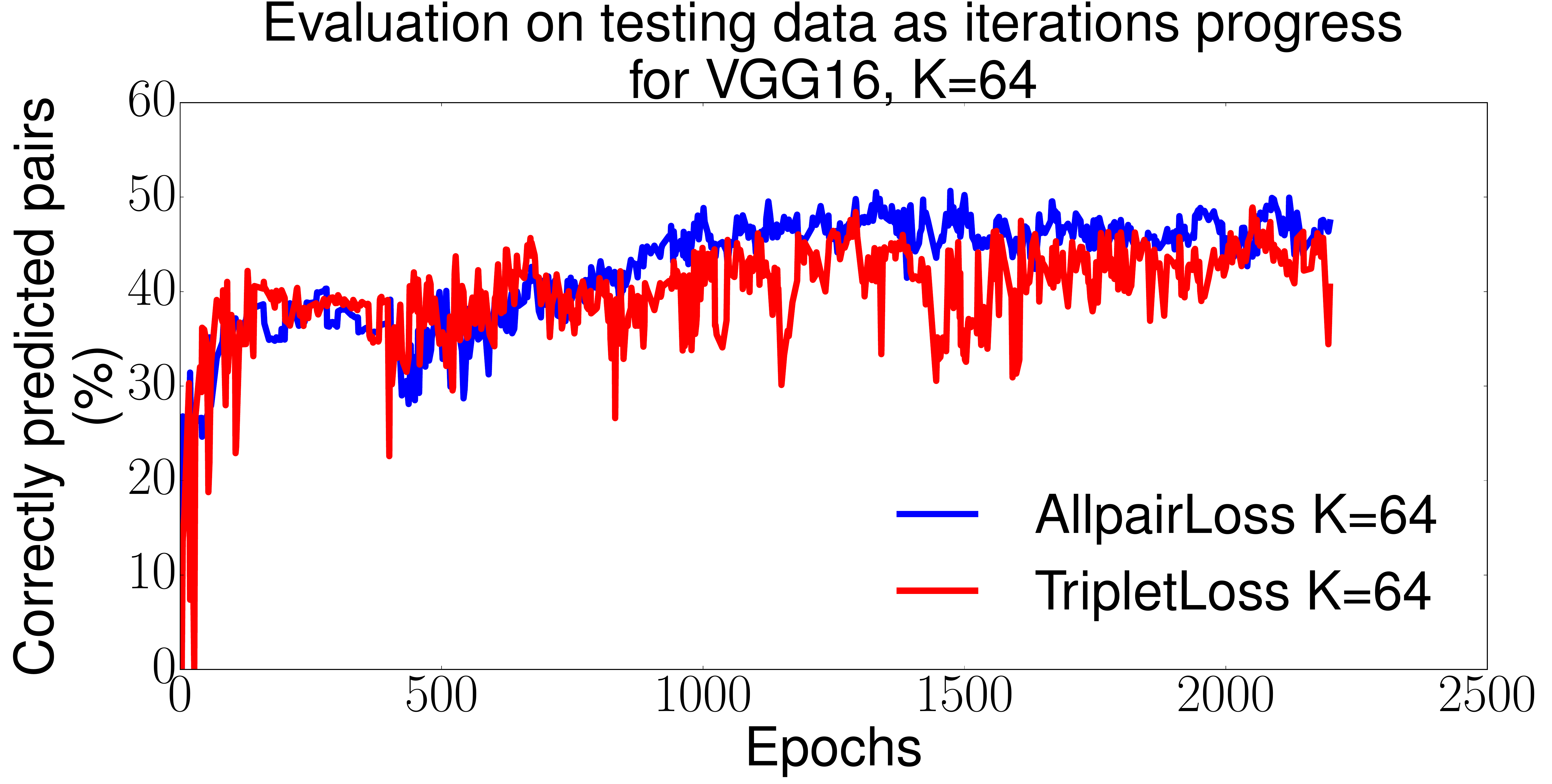}
\end{subfigure}
\caption{Effect of tripletloss vs allpairloss for VGG16 net, K=64.
(a) shows the relative training loss (ratio of loss at $i^{th}$ and $0^{th}$ iteration). (b) and (c) shows the
percentage of pairs correctly identified for training
and a separate validation data. }
\label{fig:triplet-vs-allpair-on-vgg16-k64}
\end{figure*}

\subsection{Running Times}
\label{sec:exp-flops}
Currently there is rapid progress in compute-capabilities
of GPUs. Under such circumstances it is
much more appropriate to report the number
of floating point operations (FLOPs) for the networks rather than
absolute running times in seconds (or milli-seconds).
We tabulate in Table \ref{tabular:gflops-data} the Giga-FLOPs of the networks
under various parameter settings.
For reference, the forward pass with VGG base network can be computed
in about 40-50ms and with decoupled nets in about 10-15ms for
640x480 3 channel images on Titan X (Pascal). VGG16 with $K=64$ is
the recommended configuration from Arandjelovic \textit{et al.} \cite{arandjelovic2016netvlad}, which results in a 32K-dimensional
descriptor which is reduced to 4096-dimensional by a linear
transformation. This linear transformation takes about 400 MB of
memory.

On the other hand, our proposed network which uses a network with decoupled
convolutions as the base CNN with channel squashing and $K=16$
results in 512-dimensional descriptor (4096-dimensional
if not using channel quashing).
It is
 able to run 3-4x faster than NetVLAD \cite{arandjelovic2016netvlad}
while having about 20x fewer floating point operations for
a 640x480 image and
5x fewer number of learnable parameters. See table \ref{tabular:gflops-data}.
It is worth noting that most of the computational load is in the
computation of per pixel descriptors and the NetVLAD layer itself takes
negligible computations compared to base CNN.

For training the networks, we use Intel i7-6800 CPU with Titan X (Pascal), 12GB GPU RAM.
It takes about
1 seconds per iteration for forward and backward pass. Note that one
iteration with batchsize 4, involve 52 images ($(6+6+1) \times 4$).

\begin{table}[]
\centering
\tiny
\begin{tabular}{@{}l|ccc|cc|cc@{}}
\toprule
CNN Layer            & \# L & D-Size & Model (MB) & \multicolumn{2}{c}{Fwd pass memory (MB)} & \multicolumn{2}{c}{GFLOPS}          \\ \midrule
\textbf{VGG16\_K16}  & \textbf{}     &      & \textbf{}  & \textbf{320x240}    & \textbf{640x480}   & \textbf{320x240} & \textbf{640x480} \\
block5\_pool         & 14.7M  & 8192      & 56.19      & 234.94              & 767.56             & 47.04            & 188.08           \\
block4\_pool         & 7.6M   & 8192            & 29.19      & 174.06              & 725.32             & 47.05            & 188.117          \\
block3\_pool         & 1.7M   & 4096            & 6.65       & 165.47              & 641.86             & 47.05            & 188.167          \\
\textbf{VGG16\_K64}  & \textbf{}  &         & \textbf{}  & \textbf{320x240}    & \textbf{640x480}   & \textbf{320x240} & \textbf{640x480} \\
block5\_pool         & 14.78M     & 32768         & 56.38      & 234.32              & 711.46             & 47.05            & 188.11           \\
block4\_pool         & 7.70M      & 32768        & 29.38      & 203.53              & 696.23             & 47.08            & 188.26           \\
block3\_pool         & 1.76M      & 16384        & 6.75       & 158.86              & 635.26             & 47.13            & 188.46           \\
\textbf{decoup\_K16} & \textbf{}    &       & \textbf{}  & \textbf{320x240}    & \textbf{640x480}   & \textbf{320x240} & \textbf{640x480} \\
pw13                 & 3.2M      & 8192         & 12         & 197.97              & 792.21             & 1.742            & 7.01             \\
pw10                 & 1.36M     & 8192          & 5          & 189.1               & 734.48             & 1.749            & 7.03             \\
pw7                  & 554K      & 4096         & 2          & 164.9               & 652.25             & 1.749            & 7.04             \\
\textbf{decoup\_K16\_r} & \textbf{}    &       & \textbf{}  & \textbf{320x240}    & \textbf{640x480}   & \textbf{320x240} & \textbf{640x480} \\
pw13                 & 3.5M      &  512        & 12         & 211.58              & 793.46             & 1.742            & 7.01             \\
pw10                 & 1.49M     &  512         & 5          & 193.86               & 739.73             & 1.749            & 7.03             \\
pw7                  & 686K      & 512         & 2          & 167.67               & 657.97             & 1.749            & 7.04             \\
\textbf{decoup\_K64} & \textbf{}   &        & \textbf{}  & \textbf{320x240}    & \textbf{640x480}   & \textbf{320x240} & \textbf{640x480} \\
pw13                 & 3.33M    & 32768          & 12         & 210.98              & 805.21             & 1.76             & 7.08             \\
pw10                 & 1.40M    & 32768          & 5          & 189.38              & 734.58             & 1.78             & 7.18             \\
pw7                  & 600K     & 16384          & 2          & 162.86              & 652.35             & 1.78             & 7.186            \\ \bottomrule
\end{tabular}
\caption{Tabulation of run time memory requirements,
learnable parameters (\# L), descriptor size (D-Size),
model size in Mega-bytes,
giga floating point operation (GFLOPs) for various configurations. We note that $block5\_pool$ for VGG16 network is equal in depth to $pw13$ for decouped network.
$block4\_pool$ and pw10 have equal depth; $block3\_pool$ and pw7
have equal depth.
K (eg. K16, K64)  refers to the number of clusters in NetVLAD layer.
We report data for input image size 320x240 and 640x480. We conclude that our proposed decoupled network is 20X faster computationally with an order of magnitude less number of parameters, while delivering about the same performance as the original NetVLAD. Our squashed channel network `decoup\_K16\_r` gives a descriptor size of 512 with
about 5\% additional forward pass memory and 2\% increase in parameter size with hardly noticible computation time increase.
NetVLAD \cite{arandjelovic2016netvlad} uses a whittening PCA
for reducing descriptor dimensionality which needs to store
a matrix of size 32Kx4K that takes about 400 MB.}
\label{tabular:gflops-data}
\end{table}

\subsection{Precision-recall Comparison}
\label{sec:exp-precision-recall-vpr-seq}

We evaluate the performance on the following datasets:
a) \textbf{GardensPoint} dataset, b) \textbf{CampusLoop} dataset \cite{merrill2018lightweight}, c) our \textbf{CampusConcourse} dataset.
Each of the datasets contains two sequences, `live` and `memory`.
Note that every image in live sequence has a corresponding image
in the memory sequence.
For evaluation, we load the memory sequence in the database and
compare this database with each of the images in live sequence
using a basic nearest neighbour search.
Further we also evaluate our performance for the mappilary
Berlin streetview dataset \cite{sunderhauf2015place} i)\textbf{berlin-kundamm} ii) \textbf{berlin-halenseeestrasse} and iii) \textbf{berlin-A100} as has been common amongst visual place recognition community.

We compare the proposed method with the recently proposed learning based
loop detection methods \textbf{CALC} by Merril and Huang \cite{merrill2018lightweight}.
Additionally we also compare with
\textbf{DBOW} \cite{galvez2012dbow} (Bag-of-visual words).
We evaluate with prominent approaches amongst visual place
recognition community,
\textbf{AlexNet} by Sunderhauf \textit{et al.} \cite{sunderhauf2015performance},
\textbf{LA-Net}, by Lopez-Antequera \textit{et al.} \cite{antequera2017};
 \textbf{original NetVLAD} \cite{arandjelovic2016netvlad};
Chen \textit{et al.} \cite{chen2017only}.

The main idea of this
evaluation is to gauge the recall rates and
discriminative performance of various
methods. We take the matches as correctly identified if
 match's index is within six indices of itself.
For our evaluation, we use the total number of positive matches as the length
of the sequence, since every image in the live sequence has a
correspondence in the memory-sequence. Total number of accepted matches
are those which satisfy the loop hypothesis. The precision-recall curves
are formed by sweeping through all the thresholds within the
full range of thresholds. The results presented here differ from those
presented in \cite{merrill2018lightweight} as it is not exactly clear
how recall=1 was achieved by them, how the DBoW was used to generate these results
and any heuristics, if any, was used to identify false positives.

As noted by Merril and Huang \cite{merrill2018lightweight},
and by Sunderhauf \textit{et al.} \cite{sunderhauf2018limits} superior precision-recall
does not fully prove the
superiority of a method in real loop-closure of a SLAM system.
Such factors as repeated objects in scenes,
similar looking scenes, invariance to rotation \& scale, computation time are important when considering a place recognition system for
SLAM's loop closure.
Another issue about such an evaluation is that it cannot gauge
a method's ability to return 'no matches' in case the query scene
is not found in the database. Also all these datasets are rather small
(about 80-200 frames) and we cannot evaluate the generalizability
of the scene description by each of the methods. For example
a dataset with multiple similar looking scene is needed to
throughly evaluate a method's performance. Rotation and scale variance
of the method cannot be evaluated with these datasets.
So for a better perspective of the usability of the methods
we also evaluate them on live SLAM sequences with manually marked loopclosure detections
for evaluation (details in section \ref{sec:exp-online-loop-detections}).

\begin{figure*}
\centering
\begin{subfigure}
	\centering
	\includegraphics[width=0.3\textwidth]{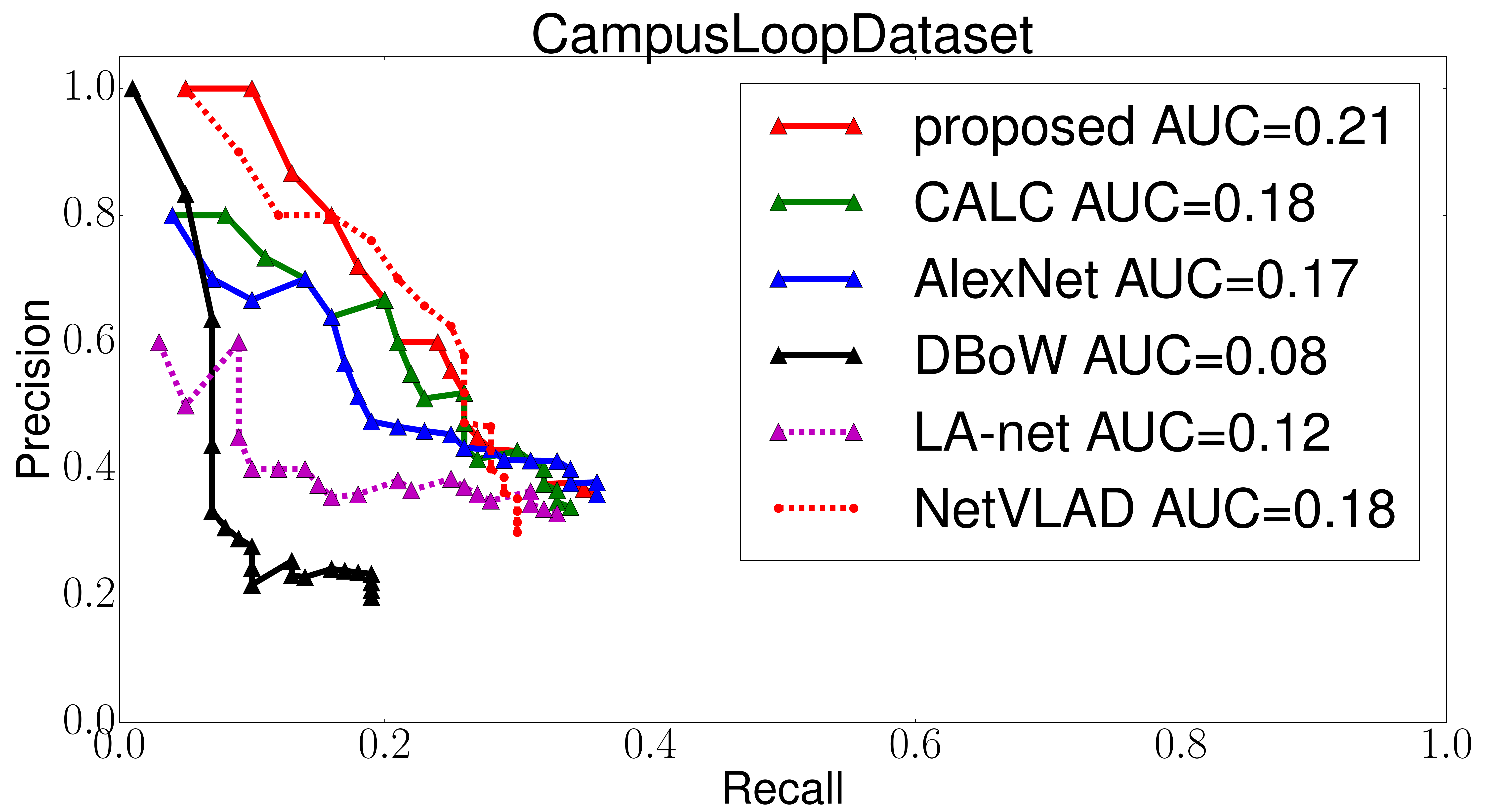}
\end{subfigure}
\begin{subfigure}
	\centering
	\includegraphics[width=0.3\textwidth]{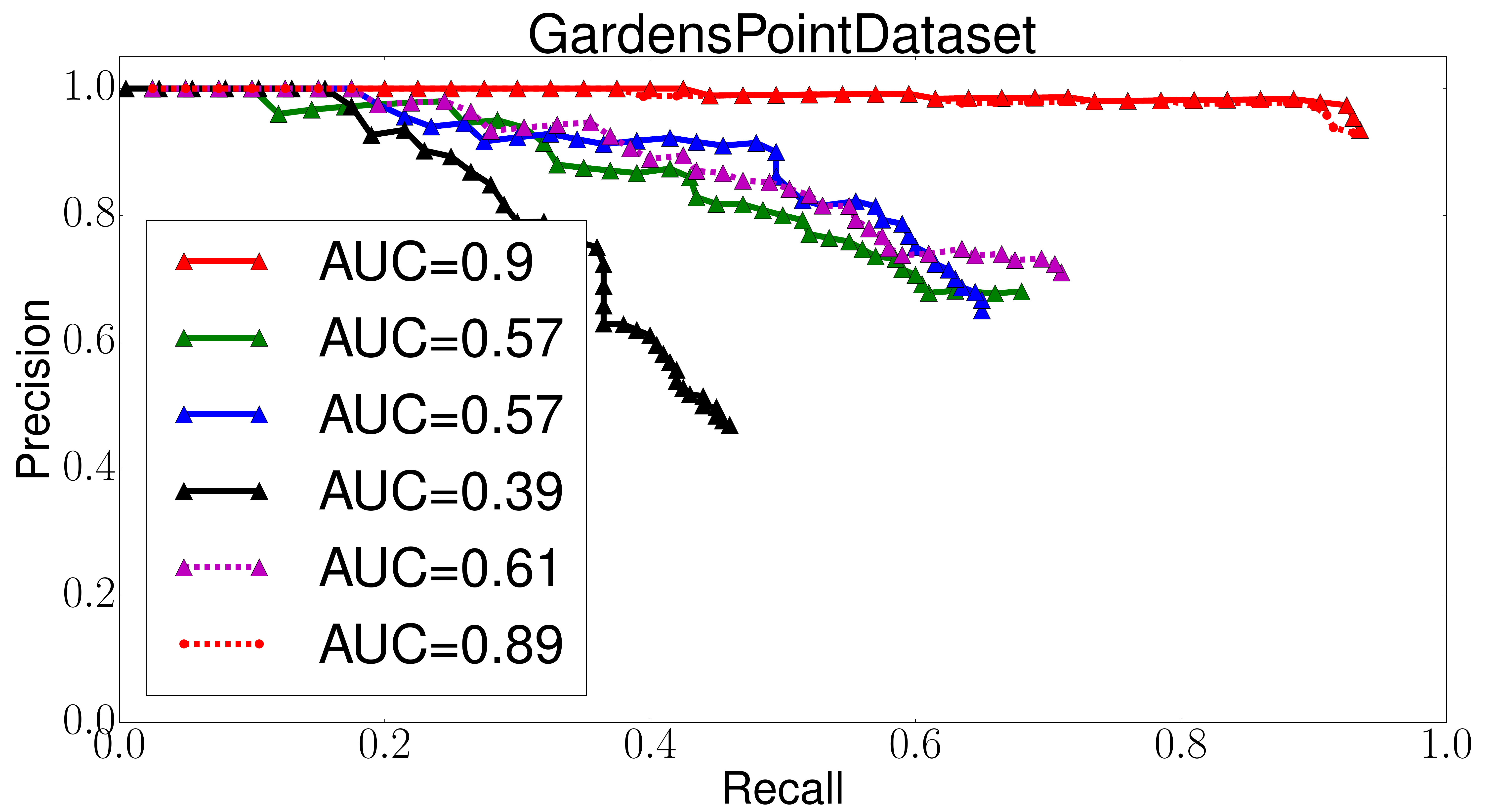}
\end{subfigure}
\begin{subfigure}
	\centering
	\includegraphics[width=0.3\textwidth]{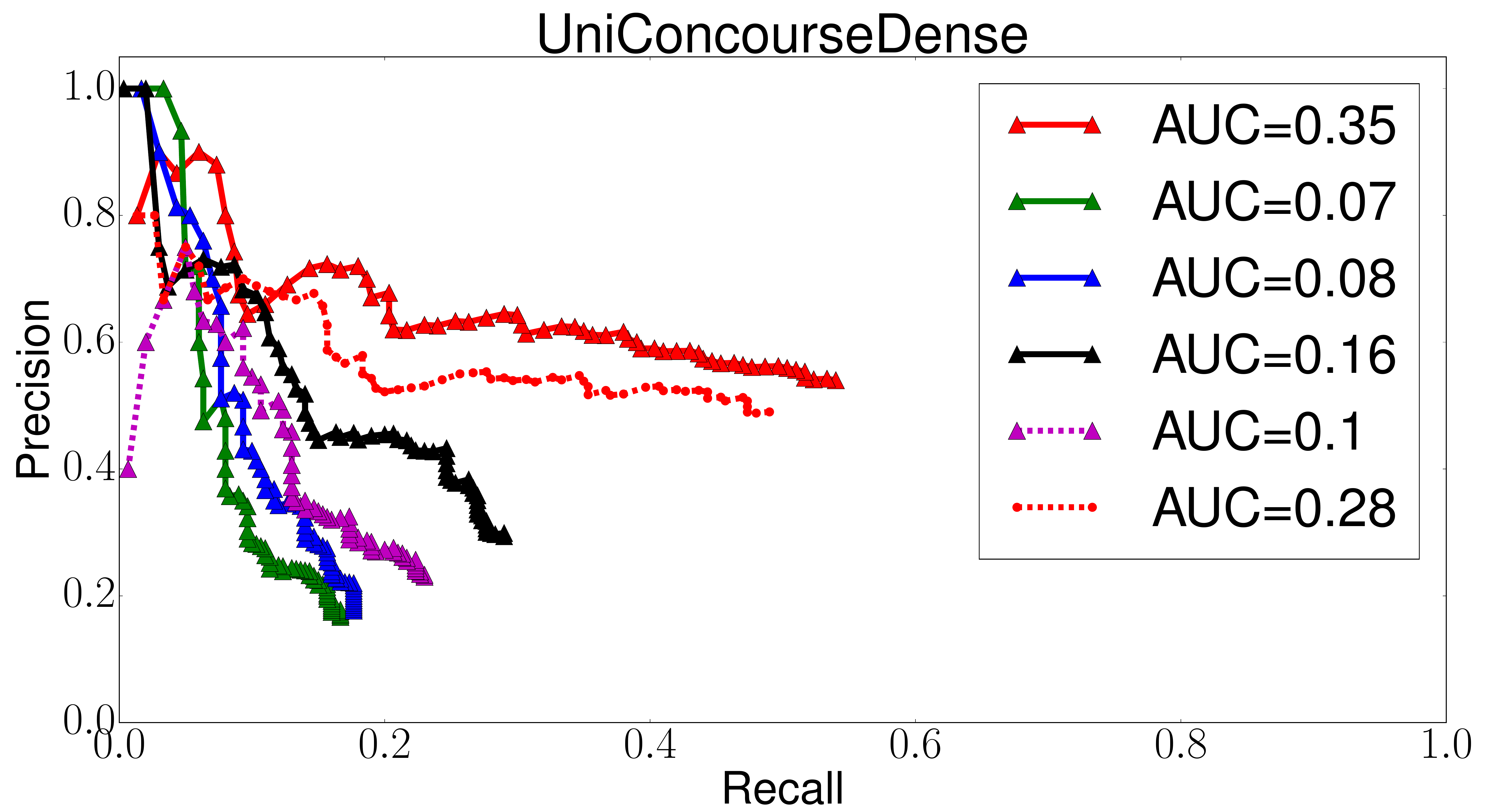}
\end{subfigure}
\caption{Precision-recall curves for various methods for loop detection.
Our method using the
decoupled net as the backend CNN gives comparable performace in CampusLoop dataset which contains
appearance changes due to snowy weather.
Our method gives a
comparable performance to the NetVLAD in other two datasets which has only large viewpoint
and in-plane rotational changes. Which is far better than
other relavant methods. }
\label{fig:precision_recall_memory_live_sequences}
\end{figure*}

\subsubsection{Walking-apart Sequence}
For precision-recall curves see Fig. \ref{fig:precision_recall_memory_live_sequences}.
The \textit{CampusLoop sequence} contains appearance variation due to changing weather
condition. Our method does not explicitly deal with this kind of variation
as it is primarily based on color cues.
The method CALC, for example, is based
on scene structure. Our method delivers comparable performance in this sequence.
For the other two testing sequences, viz.
\textit{GardenPoint} and \textit{UniConcourse}, our method performs
better than previous methods. This is attributed to the fact
that the descriptors learned by our method are able to generalize well
into identifying place revisits at
large viewpoint difference and rotational variance, which is the case in these two sequences.
When compared to  NetVLAD \cite{arandjelovic2016netvlad}
 which uses the triplet loss
and VGG network, we observe a slight boost of recall rates.
Since these sequences are very small, the higher capacity of the
proposed method  is not observable in this case.

\subsubsection{AUC Performance on Mappilary Dataset}
\begin{figure*}
\centering
\begin{subfigure}
	\centering
	\includegraphics[width=0.3\textwidth]{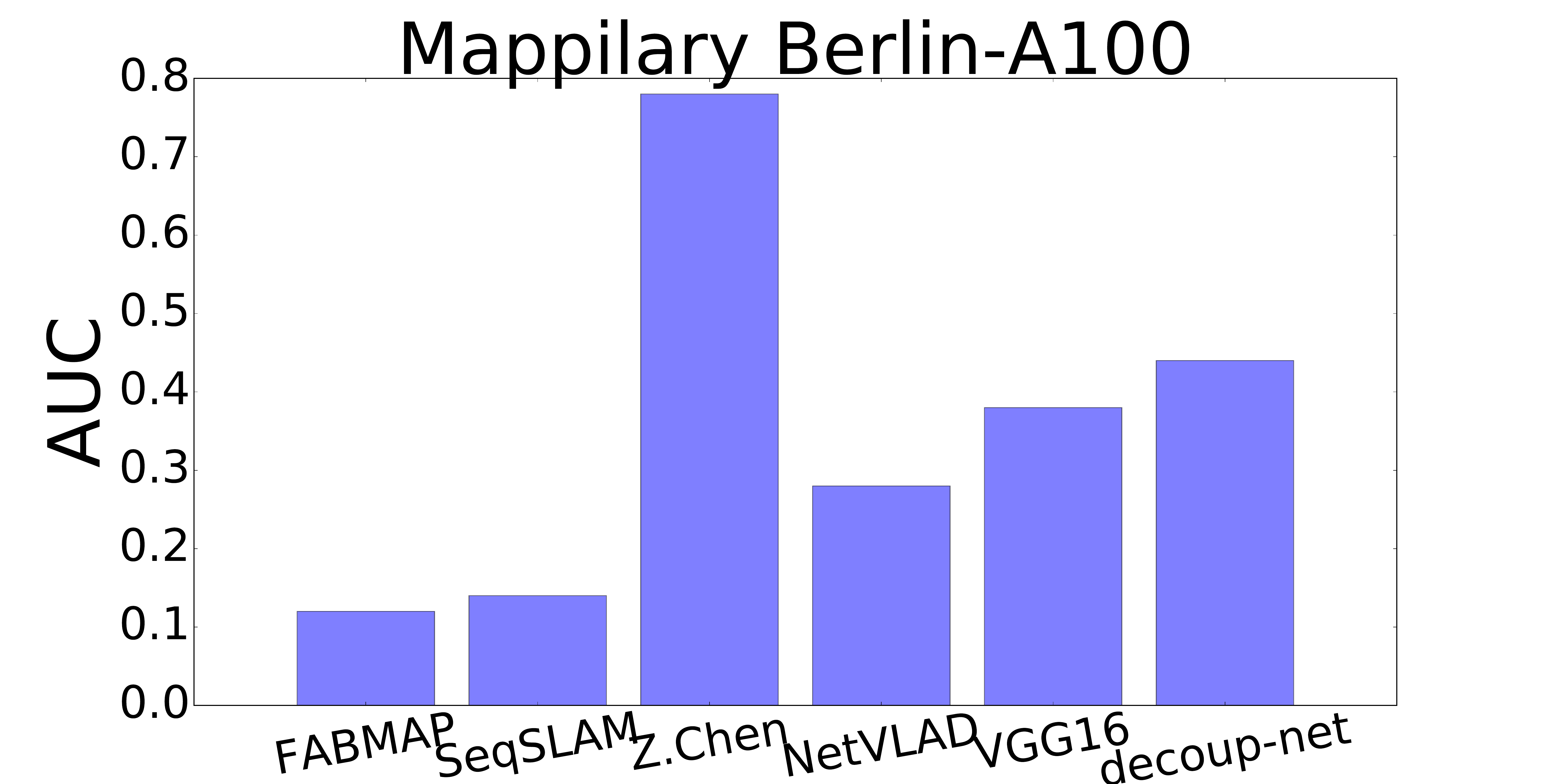}
\end{subfigure}
\begin{subfigure}
	\centering
	\includegraphics[width=0.3\textwidth]{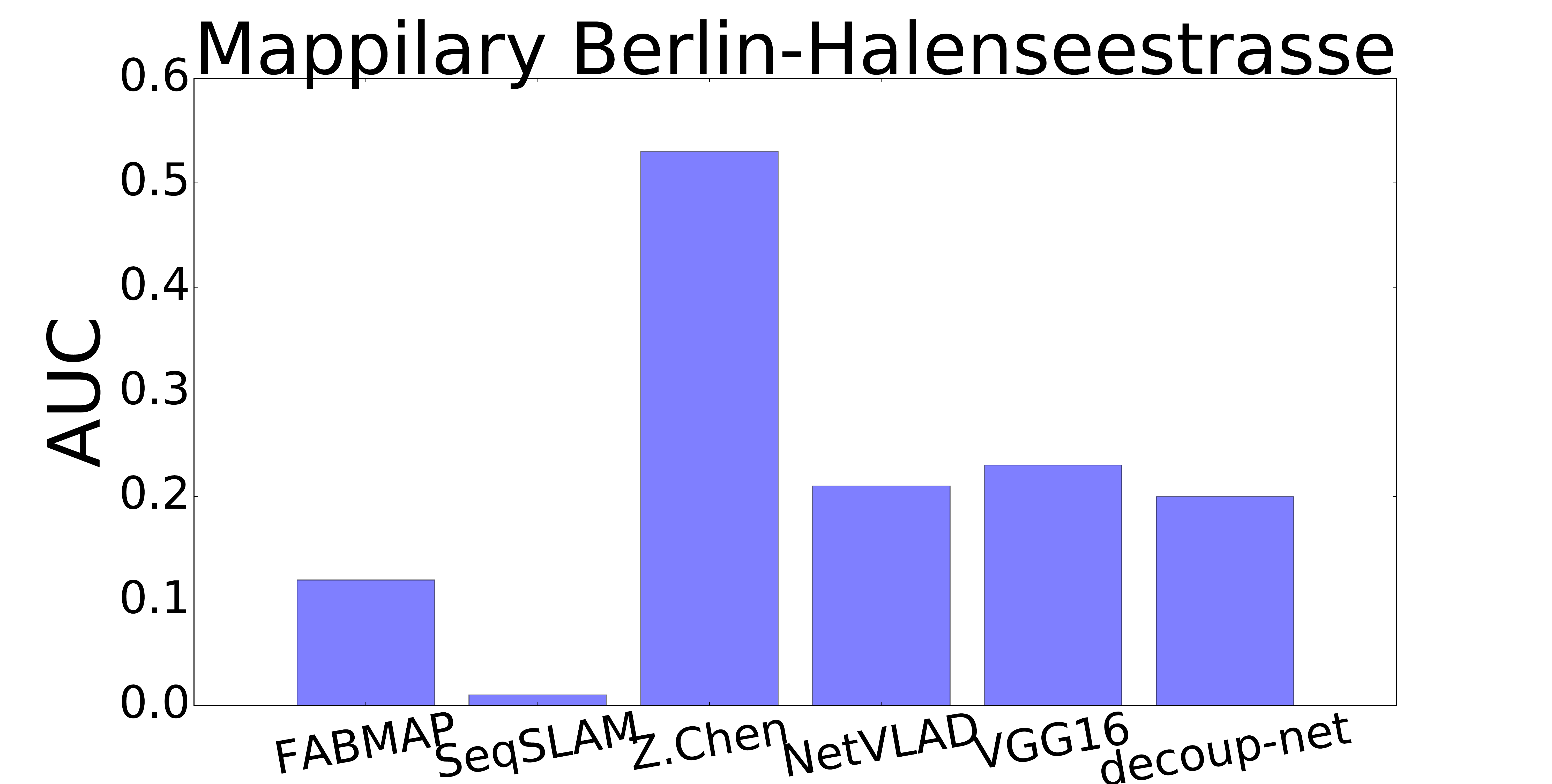}
\end{subfigure}
\begin{subfigure}
	\centering
	\includegraphics[width=0.3\textwidth]{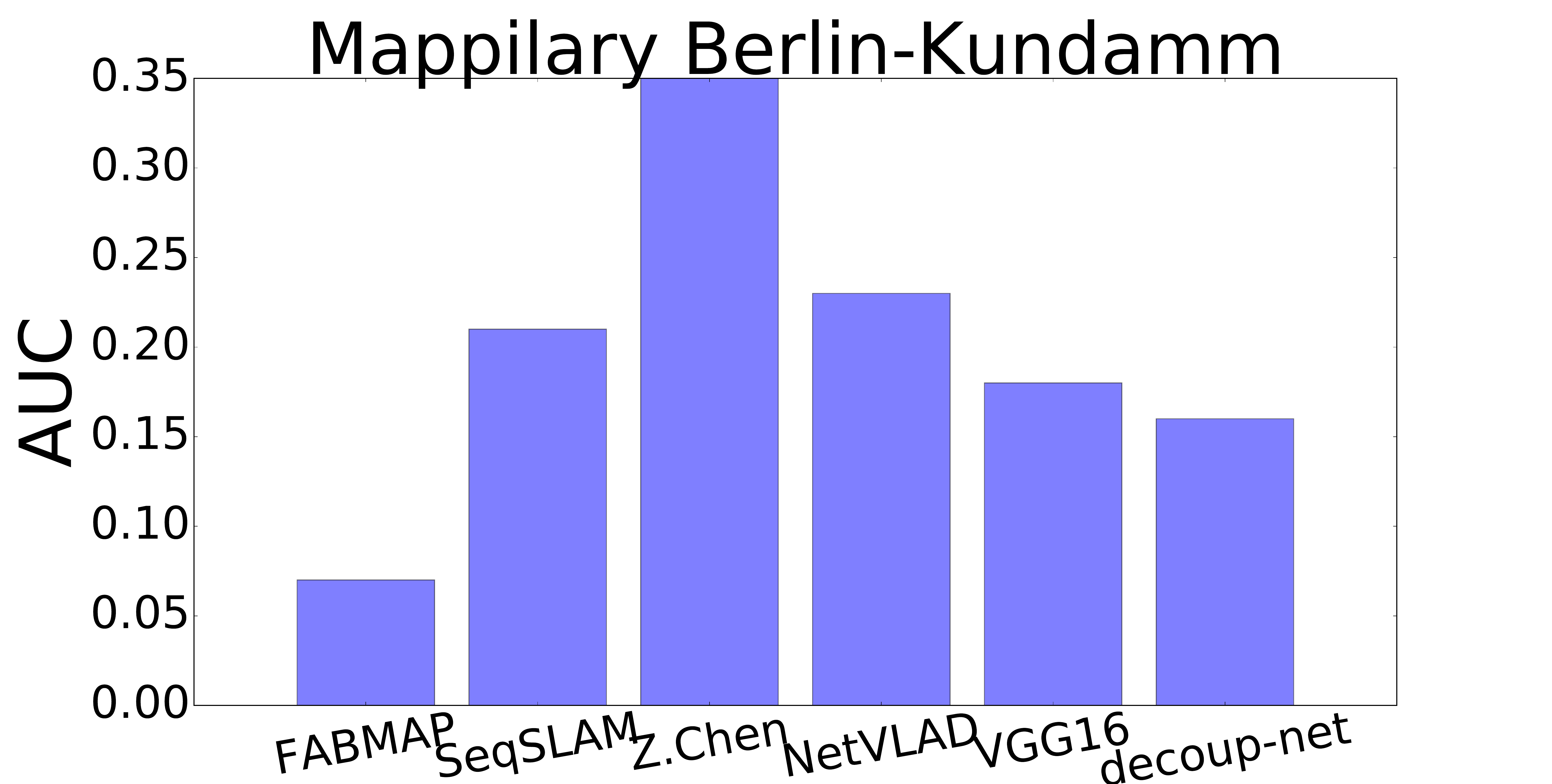}
\end{subfigure}
\caption{Comparing the methods with area under the curve (AUC) of
the precision-recall plots for the mappilary dataset. The following methods were compared: FABMAP \cite{cummins2011appearance}, SeqSLAM \cite{milford2012seqslam},
Z.Chen \cite{chen2017only}, NetVLAD \cite{arandjelovic2016netvlad}, proposed with VGG16 backend net,
proposed with decoupled net as backend net.}
\label{fig:mappilary-AUC}
\end{figure*}

We evaluate our method with other state of the art methods
with area-under-the-curve (AUC) of the precision-recall plot
on the mappilary berlin-streets dataset in Fig. \ref{fig:mappilary-AUC}. Each of the three
test sequences contain two sets of images. Note that
each image in second of the two sets has a pre-image in the
first set. These datasets are 80-200 frames each. Although
considerable viewpoint and light variation exists amongst the two
sets, there is no rotation variation. We test our method
with VGG16 backend CNN and with decoupled-net backend CNN.
We compare with FAB-MAP \cite{cummins2011appearance}, SEQSLAM \cite{milford2012seqslam},
Chen \textit{et al.} \cite{chen2017only}, NetVLAD \cite{arandjelovic2016netvlad}.
In this case, Chen \textit{et al.}'s method outperform.
It is worth noting that Chen's method \cite{chen2017only} makes use of region proposal
and is not a real-time (or near real-time) method.

\subsection{Online Loop Detections}
\label{sec:exp-online-loop-detections}

We compare the performance of the descriptors
produced from the proposed method to some of the relavant methods
with real world sequences. We introduce three sequences
and refer them as 'Live Walks Dataset', each is about 10min of
walking. The main differentiating point compared to the standard
KITTI dataset is that ours contains adversaries like revisits
under large viewpoint difference, moving objects (people),
noise, lighting changes, in-plane rotation to name a few.
Two of which
were captured with a gray scale camera and one of it was captured
with a color camera. We also provide manually marked ground truth
labels for loop detections along with odometry of the poses for visualization.
The odometry was not used for identifying loops.
Note that for a real sequence
with N keyframes there are a little less than $N^2$ pair of
loop-frames. The human was shown every pair and asked to
mark the pairs which were the same place.
Using these manual annotations,
there are 3 kinds of pairs. a) pairs not detected by
the algorithm, ie. missed pairs b) wrongly detected pairs,
ie. pairs which were in reality different places but algorithm identified it
as the same place.
c) pairs correctly identified, ie. pairs which were
marked by the algorithm as same places and  were in reality
same places.

We compare our method with some of the relavant methods on our
real world datasets. We plot the precision-recall under various
threshold settings. We define precision as the fraction
of candidate loops which were actually loops. By recall
we mean the fraction of actual loops identified. We do not use
any geometric verification step to boost our precision,
the results shown in this section are from raw image descriptor
comparison. With geometric verification,
precision of almost 100\% can be easily accomplished.

We use our method in various configurations
a) decoupled net as base CNN, K=16 (descriptor size of 4096), b) VGG16 as base CNN, K=16,
c) decoupled net with squashed channels, K=16 (descriptor size of 512). We compare with i) NetVLAD \cite{arandjelovic2016netvlad}, ii) Merril and Huang \cite{merrill2018lightweight}, iii) Sunderhauf \textit{et al.} \cite{sunderhauf2015performance}, iv) DBOW \cite{galvez2012dbow}
and v) ibow-lcd \cite{garcia2018ibow}. We acknowledge the
superior performance of Z. Chen \textit{et al.} \cite{chen2017only} method and possibly also of Sunderhauf  \textit{et al.} \cite{sunderhauf2015place}
on the mappilary dataset. However, it was not practical to
test it on our datasets which are an order of magnitude larger than
those dataset. It takes about 1-1.5 sec/frame for descriptor
computation and about 800ms-1.2 sec/pair for
descriptor comparison. For our dataset of 5000 keyframes the provided
MATLAB implementation would take almost 100 days (number of comparisons would be $5000+4999+4998+...$ ). Arguably a
faster implementation could acomplish the task in
about a day or two for a 5000 frame or 15 min walking video.
Thus this method is no where close to being real-time. Hence
it was not compared. We also note that the
running time for Sunderhauf \textit{et al.} \cite{sunderhauf2015place} is in similar range
 to Z. Chen \textit{et al.}'s method.

\begin{figure*}[!]
    \centering
    \begin{subfigure}
        \centering
        \includegraphics[width=.32\textwidth]{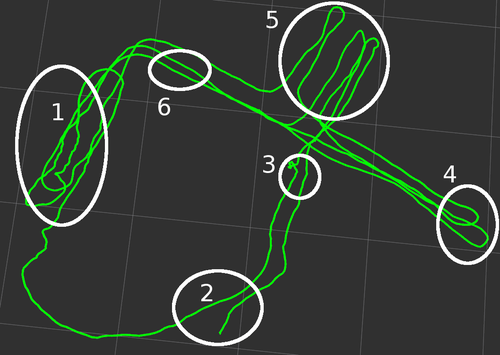}
    \end{subfigure}%
  \begin{subfigure}
        \centering
        \includegraphics[width=.32\textwidth]{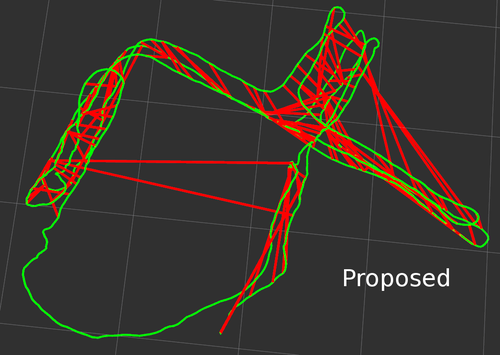}
    \end{subfigure}%
    \begin{subfigure}
        \centering
        \includegraphics[width=.32\textwidth]{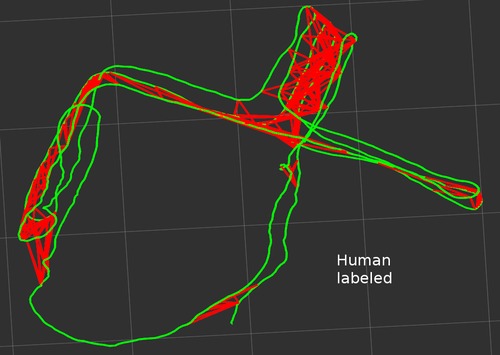}
    \end{subfigure}
    \\
    \begin{subfigure}
    \centering
    \includegraphics[width=.185\textwidth]{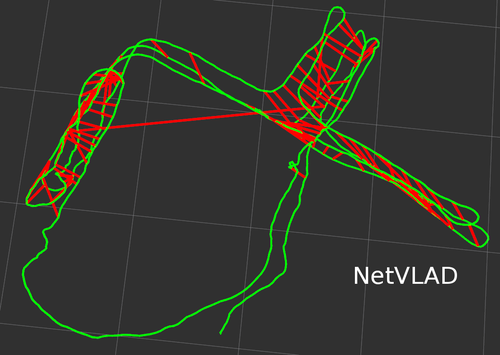}
\end{subfigure}
        \begin{subfigure}
        \centering
        \includegraphics[width=.185\textwidth]{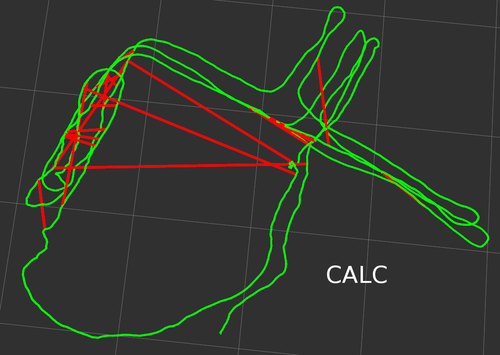}
    \end{subfigure}
    \begin{subfigure}
        \centering
        \includegraphics[width=.185\textwidth]{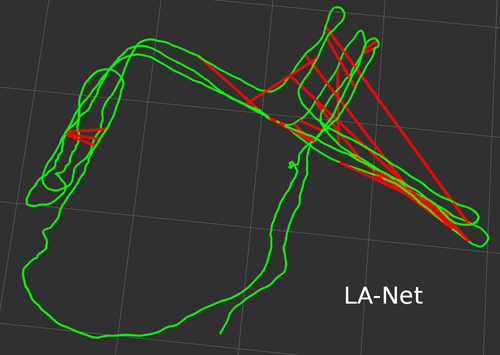}
    \end{subfigure}
     \begin{subfigure}
        \centering
        \includegraphics[width=.185\textwidth]{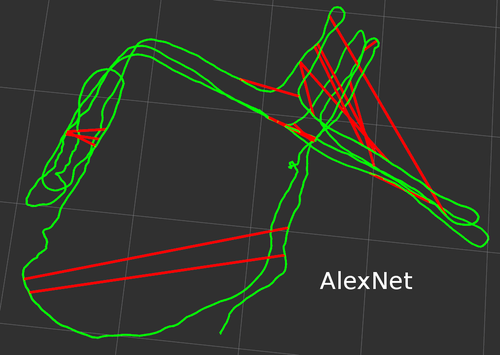}
    \end{subfigure}
    \begin{subfigure}
        \centering
        \includegraphics[width=.185\textwidth]{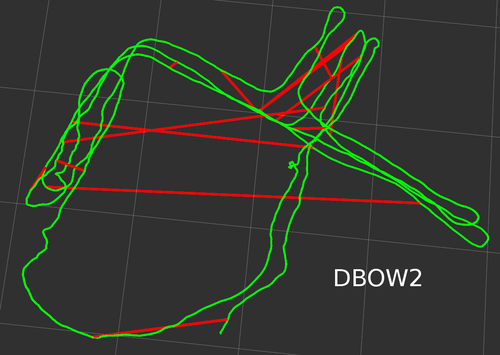}
    \end{subfigure}
    \\
        \begin{subfigure}
        \centering
        \fcolorbox{green}{white}{\includegraphics[width=0.3\textwidth]{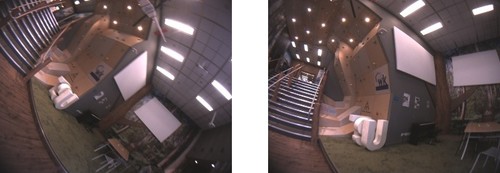}}
    \end{subfigure}
        \begin{subfigure}
        \centering
        \fcolorbox{green}{white}{\includegraphics[width=0.3\textwidth]{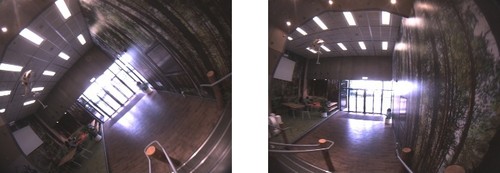}}
    \end{subfigure}
       \begin{subfigure}
        \centering
        \fcolorbox{green}{white}{\includegraphics[width=0.3\textwidth]{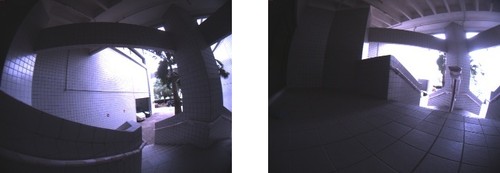}}
    \end{subfigure}
    \\
 \begin{subfigure}
        \centering
        \fcolorbox{green}{white}{\includegraphics[width=0.3\textwidth]{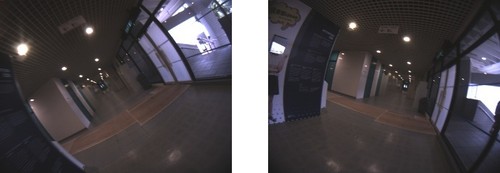}}
    \end{subfigure}
        \begin{subfigure}
        \centering
        \fcolorbox{green}{white}{\includegraphics[width=0.3\textwidth]{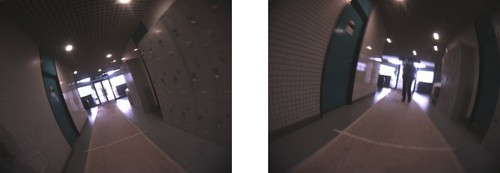}}
    \end{subfigure}
    \begin{subfigure}
        \centering
        \fcolorbox{green}{white}{\includegraphics[width=0.3\textwidth]{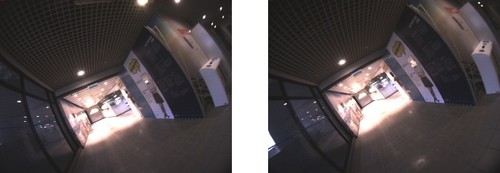}}
    \end{subfigure}
    \\

   \begin{subfigure}
        \centering
        \fcolorbox{red}{white}{\includegraphics[width=0.3\textwidth]{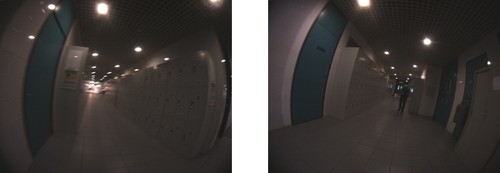}}
    \end{subfigure}
        \begin{subfigure}
        \centering
        \fcolorbox{red}{white}{\includegraphics[width=0.3\textwidth]{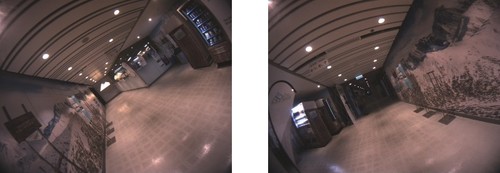}}
    \end{subfigure}
     \begin{subfigure}
        \centering
        \fcolorbox{red}{white}{\includegraphics[width=0.3\textwidth]{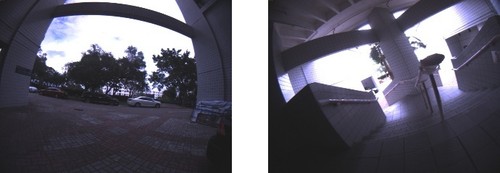}}
    \end{subfigure}
    \caption{ \textbf{Top row}: Plot of visual-inertial odometry of the sequence 'base-2';
    Loop candidates by our proposed method; human marked loop candidates;
\textbf{2nd row}: NetVLAD \cite{arandjelovic2016netvlad}; CALC \cite{merrill2018lightweight} ; LA-Net\cite{antequera2017} ; Alexnet \cite{sunderhauf2015place}   ; DBOW2 \cite{galvez2012dbow}. \textbf{Row 3 and 4}: Examples
	of correct detections by the proposed method in each of the regions. \textbf{Row 5}: Examples
	of wrong detections.
     }
    \label{fig:intro-example}
\end{figure*}

\begin{figure}[ht]
\centering
    \begin{subfigure}
        \centering
        \includegraphics[width=0.28\columnwidth]{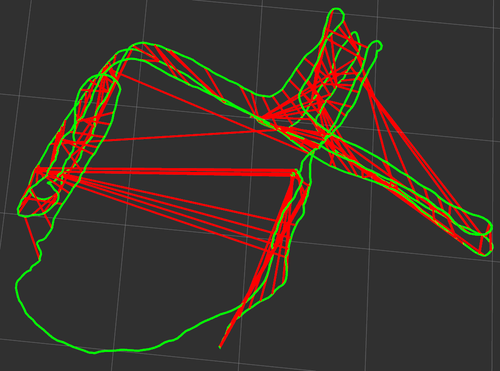}
    \end{subfigure}
        \begin{subfigure}
        \centering
        \includegraphics[width=0.28\columnwidth]{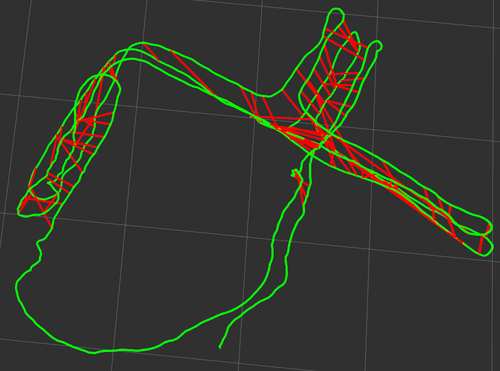}
    \end{subfigure}
        \begin{subfigure}
        \centering
        \includegraphics[width=0.28\columnwidth]{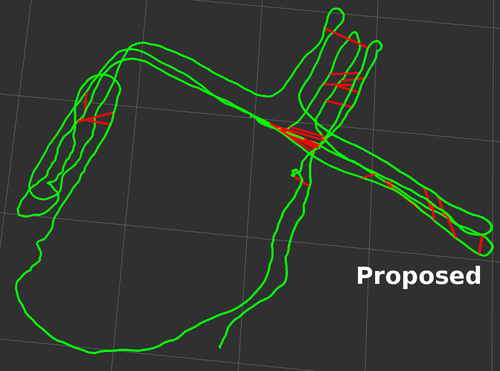}
    \end{subfigure}
    \begin{subfigure}
        \centering
        \includegraphics[width=0.28\columnwidth]{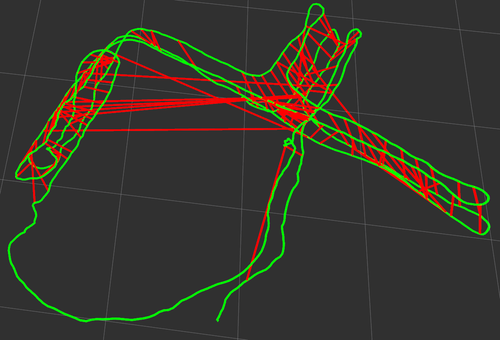}
    \end{subfigure}
        \begin{subfigure}
        \centering
        \includegraphics[width=0.28\columnwidth]{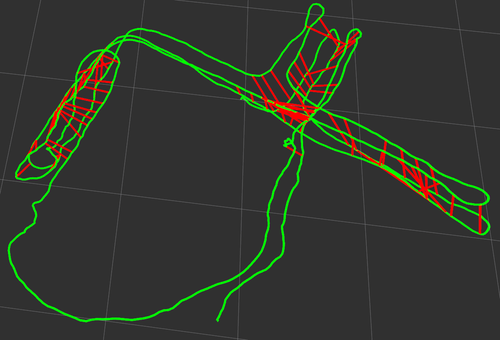}
    \end{subfigure}
        \begin{subfigure}
        \centering
        \includegraphics[width=0.28\columnwidth]{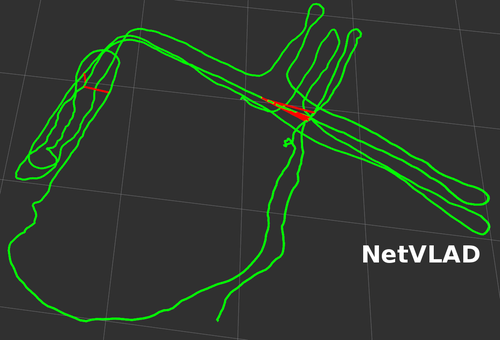}
    \end{subfigure}
    \begin{subfigure}
        \centering
        \includegraphics[width=0.28\columnwidth]{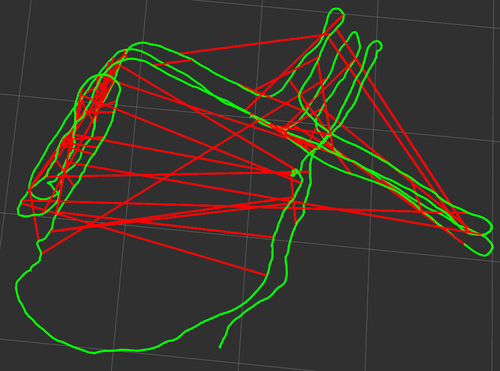}
    \end{subfigure}
        \begin{subfigure}
        \centering
        \includegraphics[width=0.28\columnwidth]{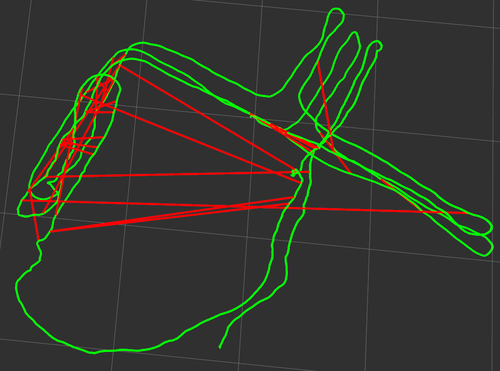}
    \end{subfigure}
        \begin{subfigure}
        \centering
        \includegraphics[width=0.28\columnwidth]{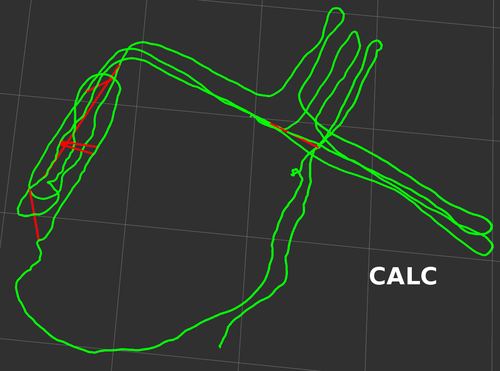}
    \end{subfigure}
    \begin{subfigure}
        \centering
        \includegraphics[width=0.28\columnwidth]{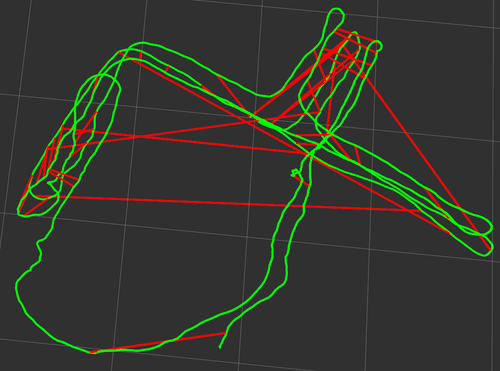}
    \end{subfigure}
        \begin{subfigure}
        \centering
        \includegraphics[width=0.28\columnwidth]{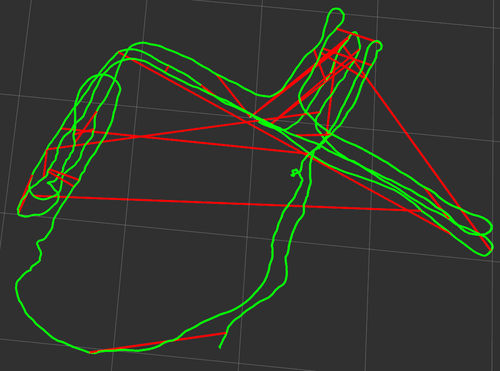}
    \end{subfigure}
        \begin{subfigure}
        \centering
        \includegraphics[width=0.28\columnwidth]{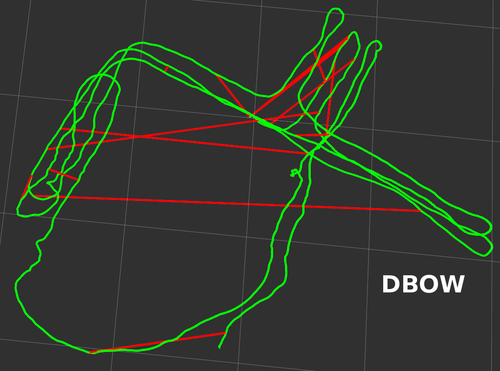}
    \end{subfigure}
    \\
    \begin{subfigure}
        \centering
        \includegraphics[width=0.90\columnwidth]{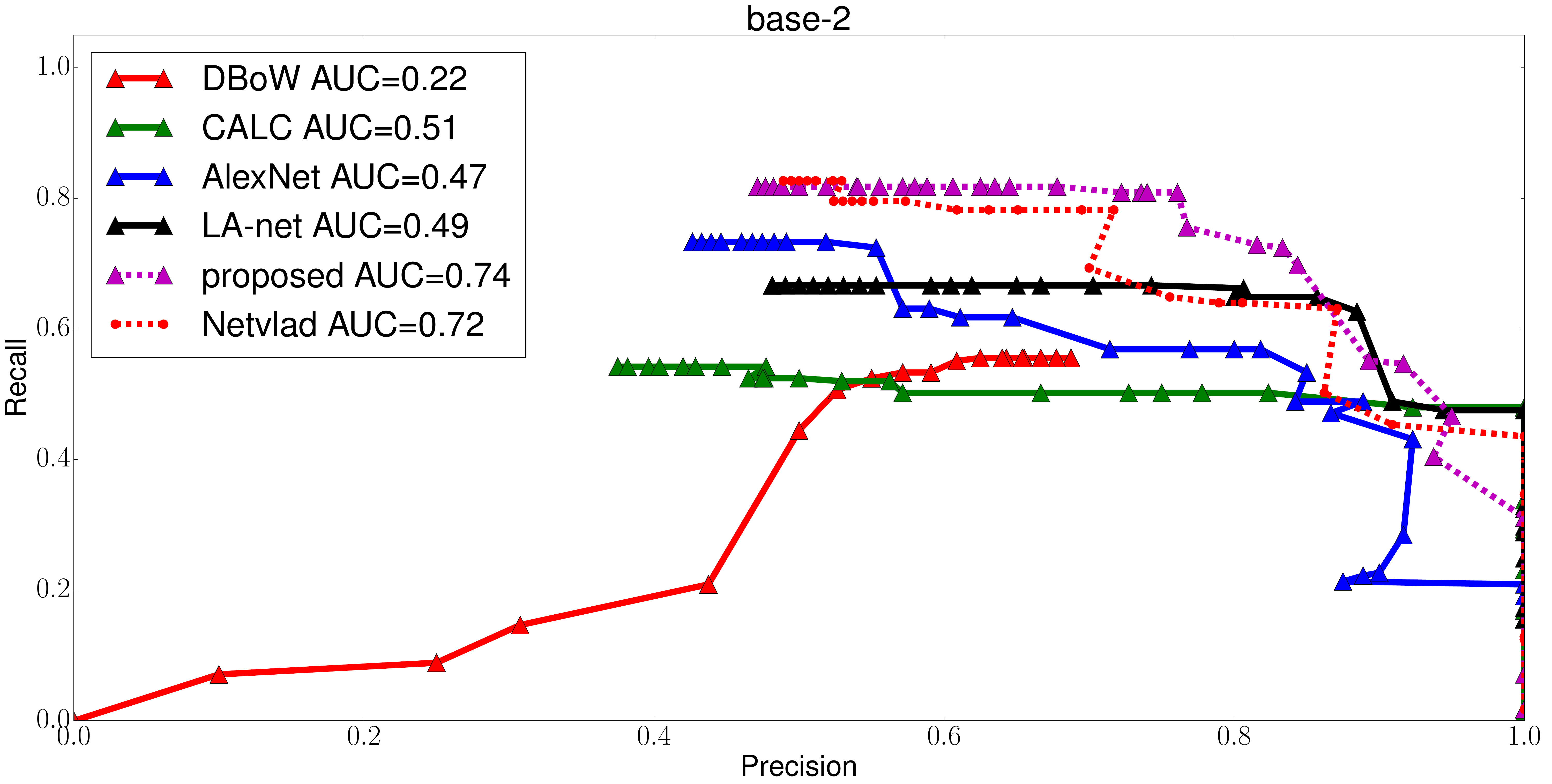}
    \end{subfigure}
\caption{Loop closure candidates (in red) as we vary the thresholds on VIO (green)
for sequence 'base-2' for the proposed method (in row-1);
NetVLAD \cite{arandjelovic2016netvlad} (in row-2);  CALC\cite{merrill2018lightweight} (in row-3) and
DBOW \cite{galvez2012dbow} (in row-4). Along the columns are various thresholds.
Leftmost is for loosest, rightmost is for tightest.
Row-5 shows the PR-curve for each method where compared
to human marked loop-candidates.}
\label{fig:seq_base_2_various_thresholds}
\end{figure}

The proposed method is able to detect revisits in all the regions for this
test sequence. We attribute this to the NetVLAD architecture which cumulates
the descriptors so as to become somewhat invariant to in-plane rotations.
Other learning based methods for example CALC totally miss
the revisits occurring under dim to moderate lighting. This can be attributed
to the fact that it is based on HOG which under dim lighting
do not provide enough spread for the histogram to generate
meaningful descriptors. CALC and DBOW as expected are able to work
in situations with in-plane rotations (in region 1 of Fig. \ref{fig:intro-example}).
AlexNet and LA-Net being essentially off-the-shelf
parameters learned for object classification task are not
invariant to rotations. DBOW however is easily confused if two scene share similar
looking textures, for example, the texture of the ceiling in otherwise
different looking scenes. DBOW also perform poorly under low-contrast scenes.
 Compared to the original NetVLAD descriptor
we observe a boost in recall rates for the proposed method.
A precision-recall curve when comparing the methods to human
marked loop candidates is presented in Fig. \ref{fig:seq_base_2_various_thresholds}.

\begin{figure}
\centering
\begin{subfigure}
        \centering
        \includegraphics[width=0.6\columnwidth]{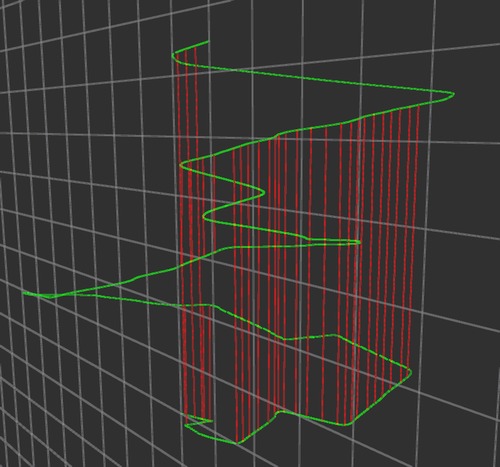}
    \end{subfigure}%
    \\
    \begin{subfigure}
        \centering
        \includegraphics[width=0.6\columnwidth]{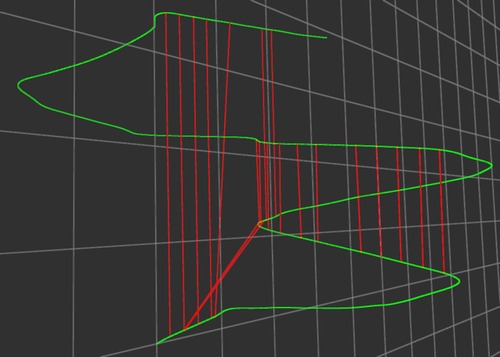}
    \end{subfigure}
\caption{The results of the proposed method on KITTI00 and KITTI05. The XY plane
is the 2d location of the trajectory. z-axis represents the frame number.
In this dataset the revisits occur at similar viewpoints, the
performance of all the compared methods is almost the same. }
\label{fig:kitti}
\end{figure}

\subsection{Full System Experiments}
We also experiment with our entire system involving relative pose computations at the
loopcandidates and pose graph solver with kidnap recovery mechanism.
Our experimental setup involves just the 'MYNT EYE D' \footnote{https://www.mynteye.com} camera.
It includes a stereo camera pair and an 200 Hz IMU with frame and IMU sync of about 1 ms.
We kidnap the camera by blocking the view of the camera and trasporting it to another location.
Additionally, we also experiment with the EuRoC MAV dataset \cite{Burri_euroc_dataset}
which also have stereo camera data and IMU.

\begin{figure}
\centering
\begin{subfigure}
        \centering
        \includegraphics[width=0.45\columnwidth]{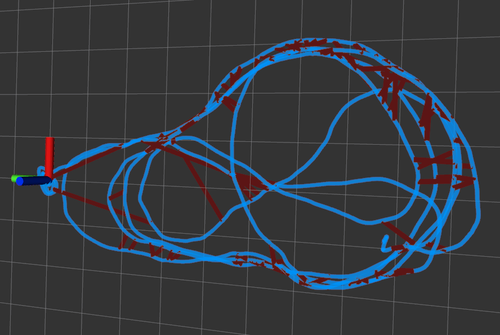}
    \end{subfigure}%
    \begin{subfigure}
        \centering
        \includegraphics[width=0.45\columnwidth]{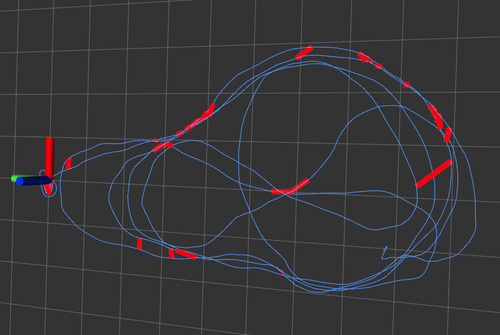}
    \end{subfigure}
    \\
    \begin{subfigure}
        \centering
        \fcolorbox{green}{white}{
        \includegraphics[width=0.9\columnwidth]{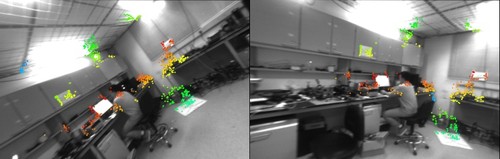}}
    \end{subfigure}
    \\
    \begin{subfigure}
        \centering
        \fcolorbox{green}{white}{
        \includegraphics[width=0.9\columnwidth]{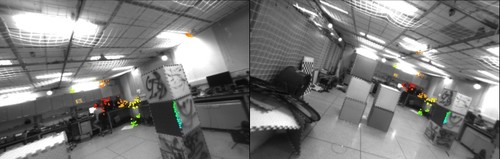}}
    \end{subfigure}
    \\
    \begin{subfigure}
        \centering
        \fcolorbox{green}{white}{
        \includegraphics[width=0.9\columnwidth]{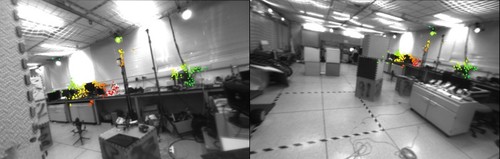}}
    \end{subfigure}
\caption{Comparing revisit detections of the proposed method (top) and
VINS-Fusion, which uses DBOW2 (2nd row). This sequence contains repeated traversal in a hall of 15mx5m at
various rotations and viewpoints. Although bag-og-words based method perform
well under fronto-parallel view it has very low recall compared to our method
on larger viewpoint difference. A side-by-side live run of this sequence is available
at \url{https://youtu.be/dbzN4mKeNTQ}. Row-3 to row-5 shows some representative looppairs which
we identified by our methods as loops but were missed out by DBOW2 in VINS-Fusion.}
\label{fig:overlay-loopedge-compare-with-vinsfusion}
\end{figure}

A representative
live real-time run of the system is shown in Fig. \ref{fig:kidnap-screenshot}.
Our system can identify and recover from kidnaps online and in realtime.
A comparison
of the loop detections with the VINSFusion system and our proposed system is shown in
Fig. \ref{fig:overlay-loopedge-compare-with-vinsfusion}. This sequence repeatedly traverse a
hall at non-fronto-parallel views and with in-plane rotations. Our system is able to
correctly recognize and compute relative poses at loopclosures involving large viewpoint differences
and in-plane rotations.

We highlight the distinguishing points of our system
compared to \textit{colmap} \cite{schoenberger2016sfm} and \textit{maplab} \cite{schneider2018maplab}.
\textit{Colmap} is a general 3D reconstruction
system and involves offline processing of unordered image sets.
The \textit{maplab} system provides an online tool, \textit{ROVIOLI} which is essentially
a visual-inertial odometry and localization front-end. Although it provides
 a console based interface for multi-session map merging, it cannot identify kidnaps
and recover from them online. Our system essentially fills in this gap.

\section{Conclusion and Future Work}
We proposed a data-driven, weakly supervised approach to learn a scene representation
for use in loopclosure module of a SLAM system. Additionally we demonstrated the use of
the disjoint dataset structure to maintain set associations of multiple coordinate systems
for online merging of multiple pose graphs.

Unstable learning was observed for the original NetVLAD  \cite{arandjelovic2016netvlad} which
made use of tripletloss for training.
This was observed to be especially prominent when trained with smaller number of
clusters. The issue was mitigated with use of the proposed allpairloss function.
This resulted in higher performance even with a smaller number of cluster in the NetVLAD layer.
For realtime performance
we made use the decoupled convolutional
layer instead of the standard convolutions for speed.
The network with decoupled convolutions are almost 3X faster
in computation time with 5-7X fewer learnable parameters.

To evaluate precision-recall performance for loopclosure detection in a real SLAM system,
we compare our
method with popular BOVW-based methods along with state-of-the-art
CNN-based methods on real world sequences.
Qualitative and quantitative experiments on standard datasets as
well as self-captured challenging sequences
with adversaries including revisits at large viewpoint difference, in-plane
rotation, dim lighting etc. suggest that proposed method can identify loopcandidates
under substantial viewpoint difference. We also observe a boost in recall
rates when compared to training with original NetVLAD. Also our
descriptors are found to be fairly invariant to rotation and lighting changes.

In addition to the precision-recall we also demonstrate the real-time working of our method
as a pluggable module for VINS-Fusion. Our system is not only able to reduce drift but also
identify and recover from complicated kidnap scenarios and failures.

A robust place recognition module is a critical element for
the SLAM dependent fields: long-term autonomy
 and augmented reality.
In addition to a robust and discriminative image
representation, the use of text and object information to further disambiguate
similar looking places and provide semantic cues to underlying planning
methods could be a way forward for a truly intelligent and scalable place recognition system.
To aid the development of such a system, we opensource
our implementation and our dataset along with human annotated looppairs
to the research community.

\bibliography{root,lit_review}

\end{document}